\documentclass[10pt,journal,compsoc]{IEEEtran}
\ifCLASSOPTIONcompsoc
  \usepackage[nocompress]{cite}
\else
  \usepackage{cite}
\fi
\ifCLASSINFOpdf
\else
\fi

\usepackage{epsfig}
\usepackage{graphicx}
\usepackage{amsmath}
\usepackage{amssymb}
\usepackage{enumitem}

\usepackage[pagebackref=true,breaklinks=true,letterpaper=true,colorlinks,bookmarks=false,urlcolor=red,citecolor=cyan]{hyperref}
\usepackage{gensymb,graphics,subfigure,inputenc}
\usepackage[linesnumbered,algo2e,boxed]{algorithm2e}
\graphicspath{{Figure/}}
\usepackage[figuresright]{rotating}
\usepackage{amsmath,amssymb,textcomp,subfigure,multirow,upgreek}
\usepackage{booktabs}
\usepackage{color,mdwlist}

\newcommand{\hao}[1]{\textcolor{black}{#1}}

\usepackage{ragged2e}

\usepackage{review}

\usepackage{graphicx,fontawesome}
\graphicspath{{Figures/}}

\usepackage{caption}
\usepackage{wrapfig}

\usepackage{enumitem}
\setitemize{noitemsep,topsep=0pt,parsep=0pt,partopsep=0pt}

\begin{document}
\title{\hao{Graph Transformer GANs with Graph Masked Modeling for Architectural Layout Generation}}
\author{Hao~Tang,
    Ling Shao,
        % Zhenyu Zhang,
	Nicu~Sebe,
	Luc Van Gool
	\IEEEcompsocitemizethanks{
	    \IEEEcompsocthanksitem Hao Tang and Luc Van Gool are with the Department of Information Technology and Electrical Engineering, ETH Zurich,  Zurich 8092, Switzerland. E-mail: hao.tang@vision.ee.ethz.ch
        \protect
        \IEEEcompsocthanksitem
        Luc Van Gool is also with INSAIT, at Sofia Un. St. Kliment Ohridski in Bulgaria. 
        \protect
	    \IEEEcompsocthanksitem Ling Shao is with the UCAS-Terminus AI Lab, University of Chinese Academy of Sciences, Beijing 100049, China. \protect
	    \IEEEcompsocthanksitem Nicu Sebe is with the Department of Information Engineering and Computer Science (DISI), University of Trento, Italy. \protect
        }% <-this % stops an unwanted space
	\thanks{Corresponding author: Hao Tang.}
}


\markboth{IEEE Transactions on Pattern Analysis and Machine Intelligence}%
{Shell \MakeLowercase{\textit{et al.}}: Bare Demo of IEEEtran.cls for Computer Society Journals}



\IEEEtitleabstractindextext{%
%\begin{abstract}
%The abstract goes here.
%\end{abstract}
\justify
\begin{abstract}
We present a novel graph Transformer generative adversarial network (GTGAN) to learn effective graph node relations in an end-to-end fashion for challenging graph-constrained \hao{architectural layout generation tasks}. The proposed graph-Transformer-based generator includes a novel graph Transformer encoder that combines graph convolutions and self-attentions in a Transformer to model both local and global interactions across connected and non-connected graph nodes. Specifically, the proposed connected node attention (CNA) and non-connected node attention (NNA) aim to capture the global relations across connected nodes and non-connected nodes in the input graph, respectively. The proposed graph modeling block (GMB) aims to exploit local vertex interactions based on a house layout topology. Moreover, we propose a new node classification-based discriminator to preserve the high-level semantic and discriminative node features for different house components. To maintain the relative spatial relationships between ground truth and predicted graphs, we also propose a novel graph-based cycle-consistency loss. Finally, we propose a novel self-guided pre-training method for graph representation learning. This approach involves simultaneous masking of nodes and edges at an elevated mask ratio (i.e., 40\%) and their subsequent reconstruction using an asymmetric graph-centric autoencoder architecture. This method markedly improves the model's learning proficiency and expediency.
Experiments on \hao{three} challenging graph-constrained \hao{architectural layout generation tasks} (i.e., \hao{house layout generation, house roof generation, and building layout generation}) with \hao{three} public datasets demonstrate the effectiveness of the proposed method in terms of objective quantitative scores and subjective visual realism. New state-of-the-art results are established by large margins on \hao{these three} tasks.
\end{abstract}

}

\maketitle

\IEEEdisplaynontitleabstractindextext

% \fi
\IEEEpeerreviewmaketitle



\section{Introduction}

\begin{figure*}[!t] \small
	\centering
	\includegraphics[width=1\linewidth]{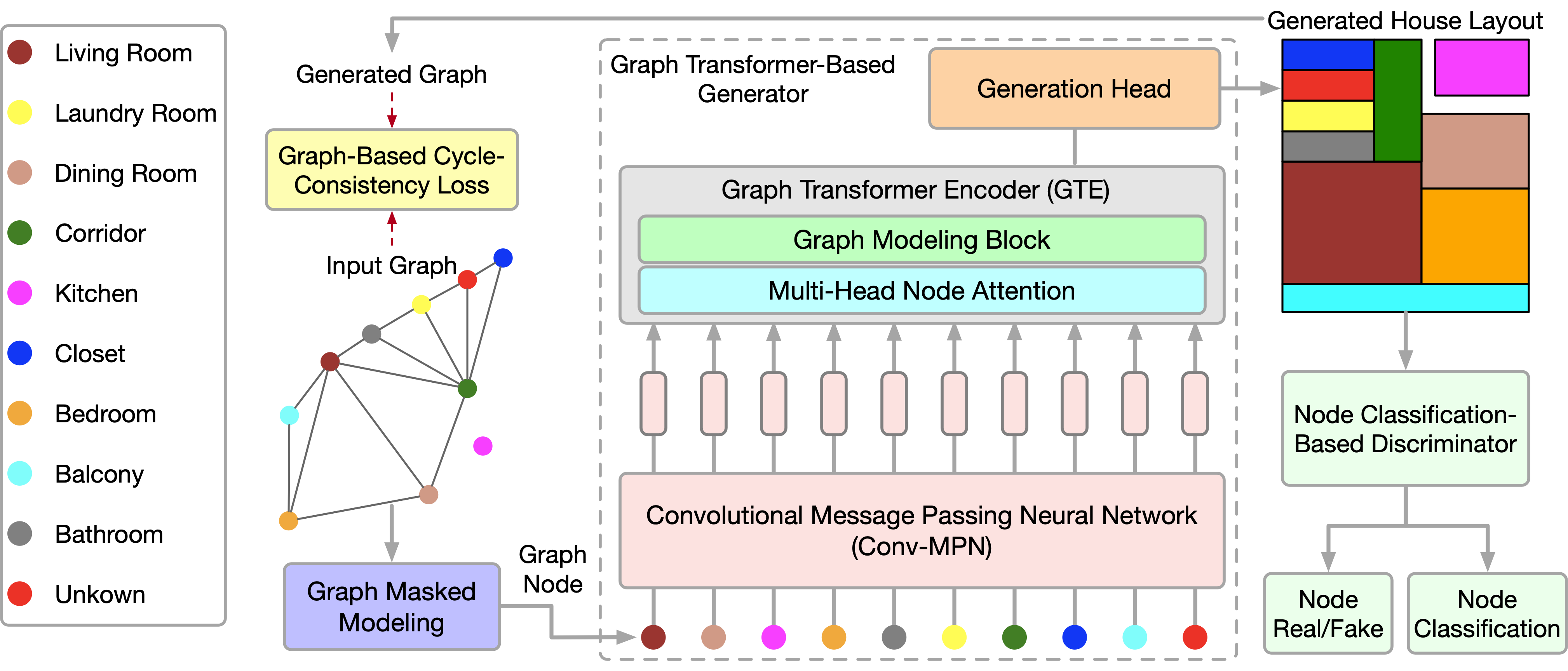}
	\caption{\hao{Overview of the proposed GTGAN++ on house layout generation. It consists of a novel graph masked modeling module, a novel graph Transformer-based generator $G$, and a novel node classification-based discriminator $D$. 
 The graph masked modeling method randomly masks 
graph nodes and edges from the input graph and recon-
structs the missing nodes and edges in the graph space, boosting the learning process of
graph representation.
		The generator takes graph nodes as input and aims to capture local and global relations across connected and non-connected nodes using the proposed graph modeling block and multi-head node attention, respectively.
		Note that we do not use position embeddings since our goal is to predict positional node information in the generated house layout.
		The discriminator $D$ aims to distinguish real and generated layouts and simultaneously classify the generated house layouts to their corresponding room types.
		The graph-based cycle-consistency loss aligns the relative spatial relationships between ground truth and predicted nodes.
}}
	\label{fig:framework}
	\vspace{-0.4cm}
\end{figure*}

\sloppy
This paper focuses on converting an input graph to a realistic house footprint, as depicted in Figure~\ref{fig:framework}.
Existing house generation methods such as \cite{wang2019planit,hu2020graph2plan,ashual2019specifying,johnson2018image,nauata2020house,wu2020pq,qian2021roof}, typically rely on building convolutional layers. 
However, convolutional architectures lack an understanding of long-range dependencies in the input graph since inherent inductive biases exist.
Several Transformer architectures \cite{vaswani2017attention,dosovitskiy2020image,zheng2020rethinking,wang2020max,wang2021transbts,carion2020end,zhu2020deformable,huang2020hand,huang2020hot,lin2020end,chen2020topological} based on the self-attention mechanism have recently been proposed to encode long-range or global relations, thus learn highly expressive feature representations.
On the other hand, graph convolution networks are good at exploiting local and neighborhood vertex correlations based on a graph topology.
Therefore, it stands to reason to combine graph convolution networks and Transformers to model local as well as global interactions for solving graph-constrained house generation.

To this end, we propose a novel graph Transformer generative adversarial network (GTGAN), which consists of two main novel components, i.e., a graph Transformer-based generator and a node classification-based discriminator (see Figure~\ref{fig:framework}).
The proposed generator aims to generate a realistic house from the input graph, which consists of three components, i.e., a convolutional message passing neural network (Conv-MPN), a graph Transformer encoder (GTE), and a generation head.
Specifically, Conv-MPN first receives graph nodes as inputs and aims to extract discriminative node features.
Next, the embedded nodes are fed to GTE, in which the long-range and global relation reasoning is performed by the connected node attention (CNA) and non-connected node attention (NNA) modules. 
Then, the output from both attention modules is fed to the proposed graph modeling block (GMB) to capture local and neighborhood relationships based on a house layout topology.
Finally, the output of GTE is fed to the generative head to produce the corresponding house layout or roof.
To the best of our knowledge, we are the first to use a graph Transformer to model local and global relations across graph nodes for solving graph-constrained house generation.

In addition, the proposed discriminator aims to distinguish
real and fake house layouts, which ensures that our generated house layouts or roofs look realistic.
At the same time, the discriminator classifies the generated rooms to their corresponding real labels, preserving the discriminative and semantic features (e.g., size and position) for different house components.
To maintain the graph-level layout, we also propose a novel graph-based cycle-consistency loss to preserve the relative spatial relationships between ground truth and predicted graphs.

We also propose a new method of pre-training graphs that considerably enhances the efficiency and potency of graph representation learning. To begin with, we establish a simple yet effective self-supervised task. This task involves masking a large portion of the nodes and edges in a graph, specifically 40\%, and then reconstructing the masked parts using a graph-based autoencoder. This type of self-supervised task, which is inherently challenging, encourages our model to delve deep into the complex structural and semantic details embedded in graphs.
Building on this, we put forward an asymmetric encoder-decoder scheme for graph pre-training (see Figure \ref{fig:graph}). The encoder, which borrows from the Transformer-style architecture, is composed of several graph modeling blocks. A graph neural network (GNN) is integrated into the attention layer, designed to extract both local and global information from the graph. The encoder operates solely on the graph subset that is visible to the model, with masked tokens being excluded.
On the other hand, the decoder's role is to reconstruct the graph using the latent representation learned, in addition to the masked tokens. While its structure is similar to the encoder, it is markedly lighter. This asymmetrical encoder-decoder structure offers significant computational advantages, significantly reducing the required computational resources and shortening the total duration of pre-training.

Overall, our contributions are summarized as follows:
\begin{itemize}[leftmargin=*]
	\item We propose a novel Transformer-based network (i.e., GTGAN) for the challenging graph-constrained house generation task. 
	To the best of our knowledge, GTGAN is the first Transformer-based framework, enabling more effective relation reasoning for composing house layouts and validating adjacency constraints.
	\item We propose a novel graph Transformer generator that combines both graph convolutional networks and Transformers to explicitly model global and local correlations across both connected and non-connected nodes simultaneously.
 	We also propose a new node classification-based discriminator to preserve high-level semantic and discriminative features for different types of rooms.
	\item We propose a novel graph-based cycle-consistency loss to guide the learning process toward accurate relative spatial distance of graph nodes.
     \item We propose a new approach to graph masked modeling, which comprehends a distribution over graphs by encapsulating conditional distributions of unmasked nodes and edges, contingent on the ones that have been observed. This method profoundly boosts the learning process of graph representation.
	\item Qualitative and quantitative experiments on \hao{three} challenging graph-constrained \hao{architectural layout generation} (i.e., \hao{house layout generation, house roof generation, and building layout generation}) with \hao{three} datasets demonstrate that the proposed method can generate better results than state-of-the-art methods, such as HouseGAN~\cite{nauata2020house} RoofGAN \cite{qian2021roof}, and \hao{BuildingGAN \cite{chang2021building}}.
\end{itemize}

\hao{Part of the material presented here appeared in the conference version \cite{tang2023graph}. This journal version extends \cite{tang2023graph} in several ways.
(1) We present a more detailed analysis of related works by including recently published works dealing with architectural layout generation tasks. We also add related methods to introduce graph masked modeling.
(2) We propose a novel graph masked modeling method for graph representation learning.
Our method masks random graph nodes and edges from the input graph and reconstructs the missing nodes and edges in the graph space.
Equipped with this proposed masked modeling method, our GTGAN proposed in \cite{tang2023graph} is upgraded to GTGAN++.
(3) We conduct extensive ablation studies to demonstrate the effectiveness of the proposed methods.
(4) We extend the quantitative and qualitative experiments by comparing our GTGAN and GTGAN++ with very recent works on three challenging tasks. We observe that the proposed methods achieve consistent and substantial gains compared with existing methods.}
\section{Related Work}

\noindent \textbf{Generative Adversarial Networks}
\cite{goodfellow2014generative} have been widely used for image generation \cite{karras2018style,shaham2019singan,tang2022local,tang2022multi}.
The vanilla GAN consists of a generator and a discriminator. 
The generator aims to synthesize photorealistic images from a noise vector, while the discriminator aims to distinguish between real and generated samples.
To create user-specific images, the conditional GAN (CGAN) \cite{mirza2014conditional} was proposed.
A CGAN combines a vanilla GAN and external information, such as class labels~\cite{choi2017stargan}, text descriptions \cite{han2017stackgan,tao2023galip,tao2022df,xu2022predict}, object keypoints~\cite{reed2016learning}, human skeletons~\cite{tang2020xinggan,tang2022bipartite,tang2021total,tang2020bipartite}, semantic maps~\cite{tang2019multi,park2019semantic,tang2020local,tang2021layout,tang2020dual}, edge maps \cite{tang2023edge,tang2023edge2}, or attention maps~\cite{mejjati2018unsupervised,tang2021attentiongan}.

This paper mainly focuses on the challenging graph-constrained generation task, which aims to transfer an input graph to a realistic \hao{architectural layout}.

\noindent \textbf{Graph-Constrained Layout Generation}
has been a focus of research recently \hao{\cite{dhamo2021graph,luo2020end,wang2019planit,hu2020graph2plan,lee2020neural,yamaguchi2021canvasvae,arroyo2021variational}}. 
For example, Wang et al. \cite{wang2019planit} presented a layout generation framework that plans an indoor scene as a relation graph and iteratively inserts a 3D model at each node.
Hu et al. \cite{hu2020graph2plan} converted a layout graph along with a building boundary into a floorplan that fulfills both the layout and boundary constraints. 
Ashual et al.~\cite{ashual2019specifying} and Johnson et al.~\cite{johnson2018image} tried to generate image layouts and synthesize realistic images from input scene graphs via GCNs. 
Nauata et al.~\cite{nauata2020house} proposed a graph-constrained generative
adversarial network, whose generator and discriminator are built upon a relational architecture.

Our innovation is a novel graph Transformer GAN, where the input constraint is
encoded into the graph structure of the proposed graph Transformer-based generator and node classification-based discriminator.
Experimental results show the effectiveness of the proposed GTGAN over all the leading methods.

\noindent \textbf{Transformers in Computer Vision.}
The Transformer was first proposed in \cite{vaswani2017attention} for machine translation
and has established state-of-the-art results in many natural language processing (NLP) tasks. 
Recently, the Vision Transformer (ViT) \cite{dosovitskiy2020image} equipped with global self-attention has achieved state-of-the-art results on the classification task.
Since then, Transformer-based approaches are efficient in many computer vision tasks including image segmentation \cite{zheng2020rethinking,wang2020max,wang2021transbts}, object detection \cite{carion2020end,zhu2020deformable,dai2022ao2,dong2023hotbev,dong2023speeddetr}, depth estimation \cite{yang2021transformer}, pose estimation \cite{li2022mhformer,lin2020end,li2023multi}, video inpainting \cite{zeng2020learning}, vision-and-language navigation \cite{chen2020topological}, video classification \cite{neimark2021video}, human reaction generation \cite{chopin2023interaction}, 3D pose transfer \cite{chen2022geometry,chen2021aniformer,chen2023lart}.

Different from these methods, in this paper, we adopt a Transformer-based network to tackle graph-constrained \hao{architectural layout generation tasks}.
However, integrating graph convolutional networks and ViTs is not trivial.
To this end, we propose a graph Transformer-based generator to capture both local and global relations across nodes in a graph.
To the best of our knowledge, the proposed GTGAN is the first Transformer-based \hao{architectural layout generation} framework.

\noindent \textbf{Graph Masked Modeling.}
Masked autoencoding has become a popular learning strategy in which a portion of the input signals is concealed and the hidden parts are predicted. It first achieved success in natural language processing with masked language modeling (MLM) \cite{kenton2019bert}, which is a fill-in-the-blank type learning task. In MLM, the model learns to represent text by accurately predicting masked words using the context provided by surrounding words. This concept was later adopted by masked image modeling (MIM) \cite{he2022masked,xie2022simmim}, which enhances representation learning by predicting the absent sections at the pixel or patch level. MIM has consistently set new performance benchmarks in a wide variety of downstream tasks.
Despite the popularity of masked autoencoding techniques in the field of language and vision research, their application in the graph domain is still relatively underexplored. However, a few recent studies have begun to bridge this gap. For example, Mahmood et al. \cite{mahmood2021masked} proposed a masked autoencoder for molecule generation, Hou et al. \cite{hou2022graphmae} developed a masked graph autoencoder for node classification and graph classification, and Tian et al. \cite{tian2023heterogeneous} proposed a heterogeneous graph masked autoencoder for node clustering and classification. These studies have shown that masked autoencoding can be an effective learning strategy for graph data, and there is potential for further research in this area.

Departing from the path of existing methodologies, we introduce two interrelated graph-based autoencoders in our proposed approach, each dedicated to masking nodes and edges respectively. The distinguishing factor of our approach is its capacity to enrich the model's expressiveness in the context of floorplan generation tasks. Through the process of masking and consequent reconstruction of nodes and edges, the model is adept at identifying and comprehending intricate structural interdependencies. Consequently, it not only captures granular details of individual nodes and edges but also assimilates holistic, graph-level information, leading to a robust representation of floorplans.
Of paramount importance is the observation that our approach is not limited to the domain of \hao{architectual layout generation} tasks. Its potential extends to the resolution of other graph-related challenges. We anticipate that our method will foster innovation and inspiration in subsequent graph-based tasks and methodologies.
\section{The Proposed Graph Transformer GAN}
\label{sec:3}

This section presents the details of the proposed GTGAN, which consists of a novel graph Transformer-based generator $G$, a node classification-based discriminator $D$, and a graph-based cycle-consistency loss. 
An illustration of the proposed GTGAN framework is shown in Figure~\ref{fig:framework}.

\subsection{Graph Transformer-Based Generator}
We only illustrate the details of our contributions on the house layout generation task for simplicity. The extension of the proposed contributions to the other two tasks is straightforward.
Take house layout generation in Figure~\ref{fig:framework} as an example. The generator $G$ receives a noise vector for each room and a bubble diagram as inputs. It then generates a house layout, where each room is represented as an axis-aligned rectangle. 
We represent each bubble diagram as a graph, where each node represents a room of a certain type, and each edge represents the spatial adjacency. 
Specifically, we generate a rectangle for each room, where two rooms with a graph edge should be spatially adjacent, while two rooms without an edge should be spatially dis-adjacent. 

\noindent \textbf{Input Graph Representation.}
Given a bubble diagram, we first generate a node for each room and initialize it with a $128$-d noise vector sampled from a normal distribution.
We then concatenate the noise vector with a $10$-d room type vector $\overrightarrow{t_r}$ ($r$ is a room index), encoded in the one-hot format.
Therefore, we can obtain a $138$-d vector $\overrightarrow{g_r}$ to represent the input bubble diagram as follows,
\begin{equation}
\begin{aligned}
\overrightarrow{g_r} \leftarrow \left \{ \mathbb{N}(0, 1)^{128}; \overrightarrow{t_r} \right \}.
\end{aligned}\label{eq:equation-factor}
\end{equation}
Note that, different from the highly successful ViT \cite{dosovitskiy2020image}, we use graph nodes as the input of the proposed graph Transformer instead of using image patches, which makes our framework very different.

\noindent \textbf{Convolutional Message Passing Neural Network.}
As indicated in HouseGAN \cite{nauata2020house}, Conv-MPN stores feature as a 3D tensor in the output design space. 
We thus apply a shared linear layer to expand $\overrightarrow{g_r}$ into a feature volume ${\rm \bf{g}}_r^{l=1}$ of size $16 {\times} 8 {\times} 8$, where $l {=} 1$ is the feature extracted from the first Conv-MPN layer, which will be upsampled twice using a transposed convolution
to become a feature volume ${\rm \bf{g}}_r^{l=3}$ of size $16 {\times} 32 {\times} 32$.

The Conv-MPN layer updates a graph of room-wise feature volumes via a convolutional message passing \cite{zhang2020conv}.
Specifically, we update ${\rm \bf{g}}_r^{l=1}$ over the following steps: 
1) We use a GTE to capture the long-range correlations across rooms that are connected in the input graph;
2) We employ another GTE to capture the long-range dependencies across non-connected rooms in the input graph;
3) We concatenate a sum-pooled feature across connected rooms in the input graph; 
4) We concatenate a sum-pooled feature across non-connected rooms; 
and 
5) We apply a convolutional neural network (CNN) on the combined feature.
This process can be formulated as follows,
\begin{equation}
	\begin{aligned}
		{\rm \bf{g}}_r^l \leftarrow {\rm CNN}\left[ {\rm \bf{g}}_r^l + {\rm GTE}\left( \underset{s \in {\rm N}(r)}{\rm Pool}  {\rm \bf{g}}_s^l, {\rm \bf{g}}_r^l \right) + 
		 \right. \\ \phantom{=\;\;}\left.  {\rm GTE}\left( \underset{s \in {\rm \overline{N}}(r)}{\rm Pool}  {\rm \bf{g}}_s^l, {\rm \bf{g}}_r^l \right);   \underset{s \in {\rm N}(r)}{\rm Pool}  {\rm \bf{g}}_s^l;   \underset{s \in {\rm \overline{N}}(r)}{\rm Pool}  {\rm \bf{g}}_s^l \right],
	\end{aligned}\label{eq:eq}
\end{equation}
where ${\rm {N}}(r)$ and ${\rm \overline{N}}(r)$ denote sets of rooms that are connected and not-connected, respectively;
``+'' and ``;'' denote pixel-wise addition and channel-wise concatenation, respectively.
We also explore two more variations (Eq.~\eqref{eq:eq1} and~\eqref{eq:eq2}) to validate  the effectiveness of  Eq.~\eqref{eq:eq} as follows,
\begin{equation}
	\begin{aligned}
		{\rm \bf{g}}_r^l  \leftarrow {\rm CNN}\left[ {\rm GTE}\left( \underset{s \in {\rm N}(r)}{\rm Pool}  {\rm \bf{g}}_s^l, {\rm \bf{g}}_r^l \right) + \right. \\ \phantom{=\;\;}\left.  {\rm GTE}\left( \underset{s \in {\rm \overline{N}}(r)}{\rm Pool}  {\rm \bf{g}}_s^l, {\rm \bf{g}}_r^l \right);   \underset{s \in {\rm N}(r)}{\rm Pool}  {\rm \bf{g}}_s^l;   \underset{s \in {\rm \overline{N}}(r)}{\rm Pool}  {\rm \bf{g}}_s^l \right].
		\label{eq:eq1}
	\end{aligned}
\end{equation}
\begin{equation}
	\begin{aligned}
		{\rm \bf{g}}_r^l  \leftarrow {\rm CNN}\left[ {\rm \bf{g}}_r^l +  {\rm GTE}\left( \underset{s \in {\rm N}(r)}{\rm Pool}  {\rm \bf{g}}_s^l, {\rm \bf{g}}_r^l \right) +  \right. \\ \phantom{=\;\;}\left. {\rm GTE}\left( \underset{s \in {\rm \overline{N}}(r)}{\rm Pool}  {\rm \bf{g}}_s^l, {\rm \bf{g}}_r^l \right)\right].
		\label{eq:eq2}
	\end{aligned}
\end{equation}

\begin{figure}[!t] \small
	\centering
	\includegraphics[width=1\linewidth]{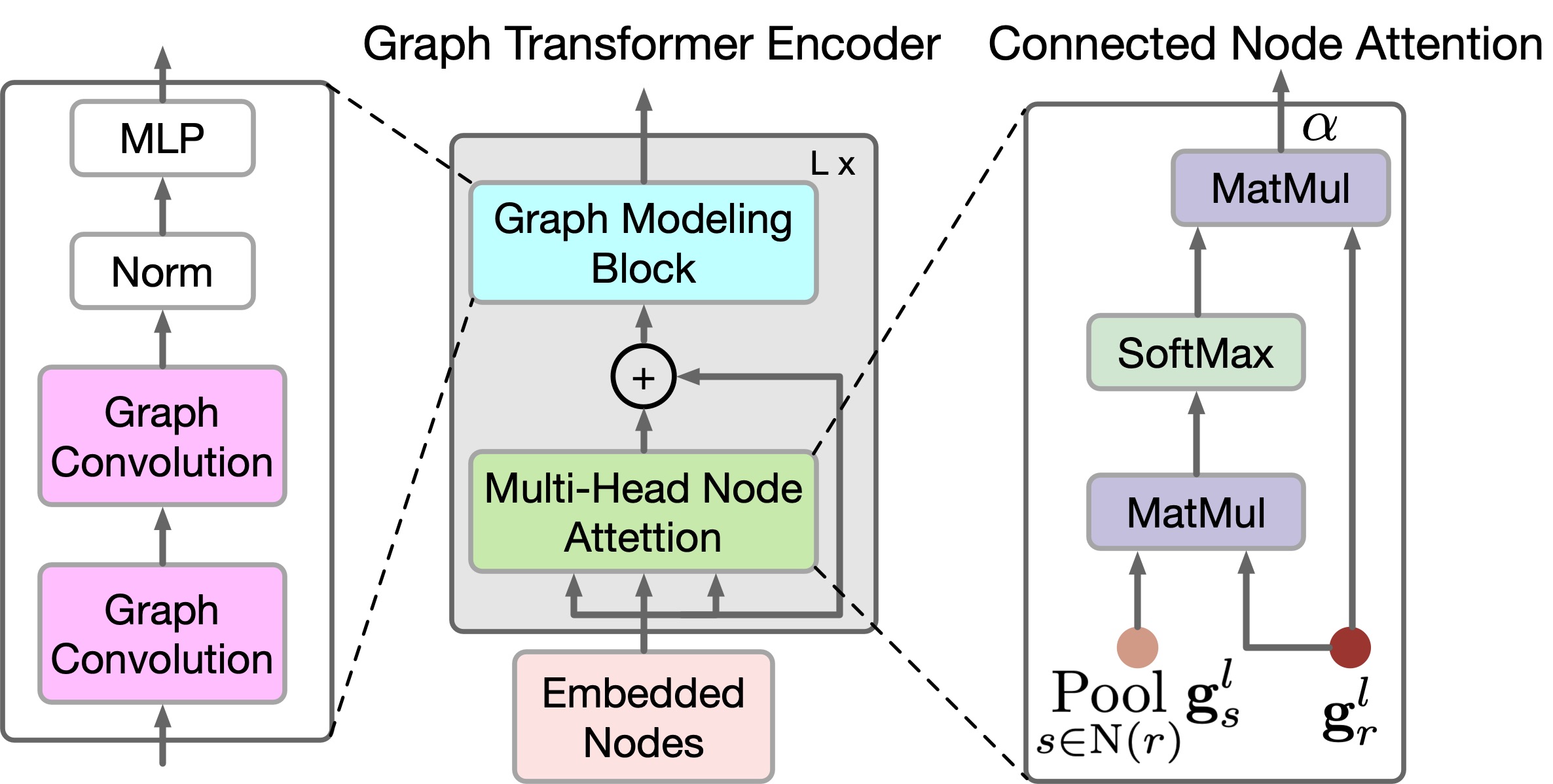}
	\caption{Overview of the proposed graph Transformer encoder, which consists of a multi-head node attention and a graph modeling block. It can capture both global and local correlations for graph-constrained house generation. This encoder consists of $L{=}8$ identical blocks. The proposed connected node attention aims to capture long-range relations across connected nodes.  Note that the proposed non-connected node attention has the same structure as the connected node attention but takes non-connected nodes as input.  It aims to capture long-range relations across non-connected nodes.}
	\label{fig:block}
	\vspace{-0.4cm}
\end{figure}

\noindent \textbf{Node Attentions in Graph Transformer Encoder.}
To capture local and global relationships across graph nodes, we propose a novel GTE, as shown in Figure~\ref{fig:block}.
GTE combines self-attention in Transformer and graph convolution networks to capture global and local correlations, respectively.
Note that we do not use position embeddings in our framework since our goal is to generate node positions in the generated house layout.

The proposed GTE is quite different from the one presented in ViT \cite{dosovitskiy2020image} since the two input modalities are different, i.e., images and graph nodes.
Thus, we extend the multi-head self-attention in \cite{dosovitskiy2020image} to the multi-head node attention, which aims to capture the global correlations across connected rooms/nodes and the global dependencies across non-connected rooms/nodes.
To this end, we propose two novel graph node attention modules, i.e., connected node attention (CNA) and non-connected node attention (NNA).
Both CNA and NNA share the same network structure.

The goal of CNA (see Figure~\ref{fig:block}) is to model the global correlations across connected rooms.
We perform a matrix multiplication between the transpose of $\underset{s \in {\rm N}(r)}{\rm Pool} {\rm \bf{g}}_s^l$ and ${\rm \bf{g}}_r^l$, and apply a Softmax function to calculate the connected node attention map $\underset{{\rm N}(r)}{\rm Att}$,
\begin{equation}
\begin{aligned}
\underset{{\rm N}(r)}{\rm Att} = {\rm softmax} \Bigg [ \frac{{\rm \bf{g}}_r^l \Big(\underset{s \in {\rm N}(r)}{\rm Pool} {\rm \bf{g}}_s^l \Big)^\top} {\sqrt{card(N(r))}} \Bigg],
\label{eq:non_local1}
\end{aligned}
\end{equation}
where $card(N(r))$ is the number of connected graph nodes in a training batch.
The connected node attention map $\underset{{\rm N}(r)}{\rm Att}$ measures the connected node's impact on other connected nodes. 
Then we perform matrix multiplication between ${\rm \bf{g}}_r^l$ and the transpose of $\underset{{\rm N}(r)}{\rm Att}$. 
Lastly, we multiply the result by a scaling parameter $\alpha$ to obtain the output,
\begin{equation}
\begin{aligned}
{\rm GTE}\left( \underset{s \in {\rm N}(r)}{\rm Pool}  {\rm \bf{g}}_s^l, {\rm \bf{g}}_r^l \right) = \alpha \sum_{1}^{N}\left (\underset{{\rm N}(r)}{\rm Att} \cdot {\rm \bf{g}}_r^l \right),
\end{aligned}
\end{equation}
where $\alpha$ is a learnable parameter, initialized to 0, and learned by the model \cite{zhang2019self}.
By doing so, each connected node in ${\rm N}(r)$ is a weighted sum of all the connected nodes.
Thus, CNA obtains a global view of the spatial graph structure and can selectively adjust rooms according to the connected attention map, improving the house layout's representations and high-level semantic consistency.

\hao{Similarly, NNA aims to capture global relations across non-connected rooms.
Specifically, we perform a matrix multiplication between the transpose of $\underset{s \in {\rm \overline{N}}(r)}{\rm Pool}  {\rm \bf{g}}_s^l$ and ${\rm \bf{g}}_r^l$, and apply a Softmax function to calculate the non-connected node attention map $\underset{{\rm \overline{N}}(r)}{\rm Att}$:
\begin{equation}
\begin{aligned}
\underset{{\rm \overline{N}}(r)}{\rm Att} = {\rm softmax} \Bigg[ \frac{{\rm \bf{g}}_r^l \Big(\underset{s \in {\rm \overline{N}}(r)}{\rm Pool} {\rm \bf{g}}_s^l \Big)^\top} {\sqrt{card(\overline{N}(r))}} \Bigg],
\label{eq:non_local2}
\end{aligned}
\end{equation}
where $card(\overline{N}(r))$ is the number of non-connected graph nodes in a training batch.
The non-connected node attention map $\underset{{\rm \overline{N}}(r)}{\rm Att}$ measures the non-connected node's impact on other non-connected nodes. 
Then we perform matrix multiplication between ${\rm \bf{g}}_r^l$ and the transpose of $\underset{{\rm \overline{N}}(r)}{\rm Att}$. 
Lastly, we multiply the result by a scale parameter $\beta$ to obtain the output as follows,
\begin{equation}
\begin{aligned}
{\rm GTE}\left( \underset{s \in {\rm \overline{N}}(r)}{\rm Pool}  {\rm \bf{g}}_s^l, {\rm \bf{g}}_r^l \right) = 
\beta \sum_{1}^{N}\left (\underset{{\rm \overline{N}}(r)}{\rm Att} \cdot {\rm \bf{g}}_r^l \right).
\end{aligned}
\end{equation}
By doing so, each non-connected node in ${\rm \overline{N}}(r)$ is a weighted sum of all the non-connected nodes.
Finally, we perform an element-wise sum with ${\rm \bf{g}}_r^l$ so that the updated node feature can capture both connected and non-connected spatial relations.
This process can be expressed as follows,
\begin{equation}
\begin{aligned}
{\rm \bf{g}}_r^l  \leftarrow {\rm \bf{g}}_r^l + {\rm GTE}\left( \underset{s \in {\rm N}(r)}{\rm Pool}  {\rm \bf{g}}_s^l, {\rm \bf{g}}_r^l \right) + {\rm GTE}\left( \underset{s \in {\rm \overline{N}}(r)}{\rm Pool}  {\rm \bf{g}}_s^l, {\rm \bf{g}}_r^l \right).
\end{aligned}
\label{eq:method2}
\end{equation}
}

\noindent \textbf{Graph Modeling in Graph Transformer Encoder.}
While CNA and NNA are useful for extracting long-range and global dependencies, it is less efficient at capturing fine-grained local information in complex house data structures. To fix this limitation, we propose a novel graph modeling block, as shown in Figure~\ref{fig:block}. 

Specifically, given the features ${\rm \bf{g}}_r^l $ generated in Eq. \eqref{eq:method2}, we further improve the local correlations by using graph convolutional networks as follows,
\begin{equation}
	\begin{aligned}
{\rm \bf{\hat{g}}}_r^l = {\rm GC}(A, {\rm \bf{g}}_r^l; P) = \sigma(A {\rm \bf{g}}_r^l P),
	\end{aligned}
	\label{eq:graph}
\end{equation}
where $A$ denotes the adjacency matrix of a graph, ${\rm GC}(\cdot)$ represents graph convolution, and  $P$ the trainable parameters. 
$\sigma(\cdot)$ is the gaussian error linear unit (GeLU) proposed in \cite{hendrycks2016gaussian} activation function that aims to provide the network non-linearity.
We follow the structure design in GraphCMR \cite{kolotouros2019convolutional} to build our graph modeling block, which can explicitly encode the graph-constrained house structure within the network and thereby improve spatial locality in the feature representations.

\noindent \textbf{Generation Head.} 
We adopt three CNN layers to convert a feature volume into a room segmentation mask of size $1 {\times} 32 {\times} 32$. 
The numbers of convolutional channels are 256, 128, and 1, respectively.
We pass the graph of segmentation masks to the proposed discriminator $D$ during the training stage. 
Finally, we fit the tightest axis-aligned rectangle for each room to generate the house layout.

\subsection{Node Classification-Based Discriminator}
The input of the proposed discriminator is a graph of room segmentation masks, either from the generator or a real one.
The segmentation masks are of size $1 {\times} 32 {\times} 32$. 
We also take a $10$-d room type vector to preserve the room type information, and then we apply a linear layer to expand it to $8192$-d.
Next, we reshape it to a tensor of size $8 {\times} 32 {\times}32$.
Thus, we use a shared three-layer CNN to convert it to a feature of size $16 {\times} 32 {\times}32$, followed by two rounds of Conv-MPN and downsampling. 
Lastly, we use another three-layer CNN to convert each room feature into a $128$-d vector $\overrightarrow{d_r}$. 
To classify ground-truth samples from the generated ones, we sum-pool over all the room vectors and then apply a single linear layer to produce a scalar $\rm {\bf{\tilde{d}}_1}$, which can be expressed as follows,
\begin{equation}
\begin{aligned}
{\rm \bf{\tilde{d}}_1}  \leftarrow {\rm Linear} \left( \underset{r}{{\rm Pool}}~\overrightarrow{d_r} \right).
\end{aligned}\label{eq:eq3}
\end{equation}
Moreover, we observe that HouseGAN \cite{nauata2020house} cannot produce very discriminative rooms, leading to similar generation results for different types of rooms.
To provide a more diverse generation for different rooms, we propose a novel node classification loss to learn more discriminative class-specific node representations.
Specifically, we sum pool over all the room vectors and add another single linear layer to output a $10$-d one-hot vector $\rm \bf{\tilde{d}}_2$, classifying generated rooms to the corresponding room labels,
\begin{equation}
\begin{aligned}
{\rm \bf{\tilde{d}}_2}  \leftarrow {\rm Linear} \left( \underset{r}{{\rm Pool}}~\overrightarrow{d_r} \right).
\end{aligned}\label{eq:eq4}
\end{equation}
We use the binary cross-entropy loss between the real room label $\rm \bf{d}_2$ and the predicted label $\rm \bf{\tilde{d}}_2$.

\begin{figure*}[!t] \small
	\centering
	\includegraphics[width=1\linewidth]{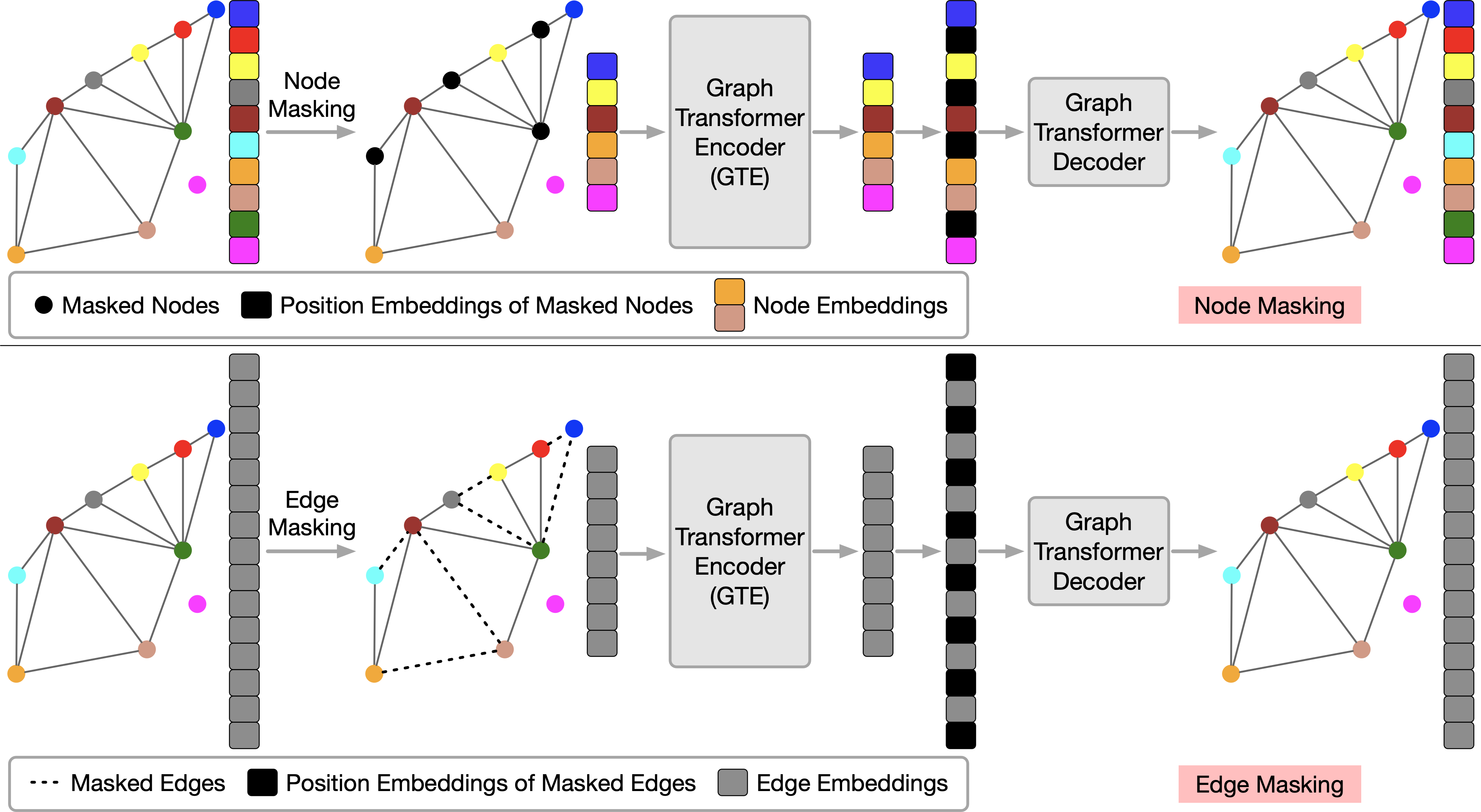}
	\caption{
 This figure reveals the intricately architected self-supervised tasks embedded within GTGAN. A substantial majority of nodes or edges are randomly obscured, following which, the GTGAN undergoes a pre-training stage. During this stage, we aim to recreate the original rooms utilizing both the latent representations and masked tokens.}
	\label{fig:graph}
	\vspace{-0.4cm}
\end{figure*}

\subsection{Graph-Based Cycle-Consistency Loss}
Providing global graph node relationship information is helpful in generating more accurate house layouts. 
To differentiate this process, we propose a novel loss based on an adjacency matrix that matches the spatial relationships between ground truth and generated graphs, as shown in Figure \ref{fig:framework}. 
Precisely, the graphs capture the adjacency relationships between each node of different rooms, and then we enforce the matching between the ground truth and generated graphs through the proposed graph-based cycle-consistency loss. 
Formally, we represent the graphs using two (square) weighted adjacency matrices of size $M{\times}M$,
\begin{equation}
	\begin{aligned}
		{\rm \textbf{G}}^{gt} & = \{g_{i,j}^{gt}\}_{\substack{i=1,\dots,M \\ j=1,\dots,M}},  \\
		{\rm \textbf{G}}^{gen} & = \{g_{i,j}^{gen}\}_{\substack{i=1,\dots,M \\ j=1,\dots,M}}.
	\end{aligned}\label{eq:eq5}
\end{equation}
The matrix ${\rm \textbf{G}}^{gt}$ contains the adjacency information computed on ground
truth graph, while ${\rm \textbf{G}}^{gen}$ has the same information computed on the
generated graph. 
Note that we adapt the network in \cite{yu2019layout} to obtain the $G^{gen}$ from the generated house layout, followed by a fully-connected layer with the size of $M{\times}M$, then reshaped it to a square matrix.

In this way, each element of the matrices provides a measure of how close the two nodes $i$ and $j$ are in the ground truth and the generated graph, respectively. 
To measure the closeness between nodes, which is a hint of the strength of the connection between them, we consider weighted matrices where each entry $g_{i,j}$ depends on the shortest distance between them. 
For example, in the ground truth graph in Figure \ref{fig:framework}, the shortest distance between the dining room and the living room is 1, the shortest distance between the dining room and the closet is 2, and the shortest distance between the bedroom and the closet is 3.
Note that we do not consider self-connections, thus $g_{i,i}^{gt} {=} g_{i,i}^{gen} {=}0$ for $i{=}1,\dots,M$. 
Moreover, non-adjacent nodes have $-1$ as the entry. 
Then, we define the proposed graph-based cycle-consistency loss as the Frobenius norm between the two adjacency matrices,
\begin{equation}
  \mathcal{L}_{gcyc} = || {\rm \textbf{G}}^{gt} - {\rm \textbf{G}}^{gen} ||_F =|| {\rm \textbf{G}}^{gt} - G({\rm \textbf{G}}^{gt}) ||_F,
 \label{eq:eq6}
\end{equation}
where $G$ is the proposed graph Transformer-based generator. 
This loss function aims to faithfully maintain the reciprocal relationships between nodes. On the one hand, disjoint parts are enforced to be predicted as disjoint. 
On the other hand, neighboring nodes are enforced to be predicted as neighboring and to match the proximity ratios.

\subsection{Implementation Details}
\noindent \textbf{Optimization Objective.}
The optimization objective of the proposed GTGAN can be written as follows,
\begin{equation}
	\begin{aligned}
		\mathcal{L} = \mathcal{L}_{gan} + \lambda_1 \cdot \mathcal{L}_{class}({\rm \bf{d}_2}, \rm {\bf \tilde{d}}_2) +  \lambda_2 \cdot   \mathcal{L}_{gcyc}, 
		\label{eq:loss} 
	\end{aligned}
\end{equation}
where  $\mathcal{L}_{gan}$, $\mathcal{L}_{class}$, and $\mathcal{L}_{gcyc}$ denote the adversarial loss, the graph node classification loss, and the graph-based cycle-consistency loss, respectively.

\noindent \textbf{Training Details.} For a fair comparison, we follow HouseGAN \cite{nauata2020house} and use the Adam solver \cite{kingma2014adam} with $\beta_1 {=} 0.5$ and $\beta_2 {=} 0.999$ as our optimizer.
The learning rates of our generator $G$ and discriminator $D$ are both set to 0.0001. 
We set the batch size to 32 on all the subsets for training. 
We also use the WGAN-GP loss as in \cite{nauata2020house} to train the network.

\section{The Proposed Graph Masked Modeling}

In this section, we delve deeper into our approach for both the pre-training stage and the ensuing fine-tuning steps. To initiate, we present an overview of the architecture. Figure \ref{fig:graph} demonstrates our proposed model which consists of two distinct branches: one focused on graph node masking and the other centered around graph edge masking. The function of each branch is to extract embeddings from the nodes or edges of the input graph, respectively. These embeddings are later utilized during the fine-tuning phase for various downstream tasks.
Our model's design has been inspired by MAE \cite{he2022masked} and BatmanNet \cite{wang2024batmannet}, and we have crafted an asymmetric encoder-decoder structure for each branch in the style of a Transformer.
The pre-training phase involves the application of our unique graph masking strategy. The graph Transformer encoder operates on partially observable house graph signals during this phase and transforms them into latent representations of nodes or edges. Subsequently, these latent representations, along with the mask tokens, are used by the graph Transformer decoder to reconstruct the original house graph.

\subsection{Pre-Training Strategy}

The methodology we've devised functions at both individual node and comprehensive graph levels. This approach equips the model with the ability to discern and interpret not only localized attributes but also the overarching and intricate information inherent in the graph.
Drawing upon principles from MAE, we design a novel task involving graph masking and reconstruction to pre-train on graph structures. As an example, when we apply our method to a house graph, we advocate for a high proportion of nodes to be randomly masked or omitted in the node branch. This mechanism forces the encoder to focus only on the unmasked nodes that persist, thus challenging it to capture the essential characteristics of these nodes.
In a similar vein, we also implement a masking technique for the edge branch. However, in this case, the input is derived from the complementary edge graph of the original graph. This method pushes the model to understand the relationships between nodes, strengthening its ability to capture complex patterns and connections within the graph.

We propose that our novel graph masking and rebuilding technique allows the pre-processed node and edge embeddings to grasp the details of a house both at the individual node/edge level and at the holistic graph level. The first facet of our approach involves significantly obscuring nodes and edges, such as at a ratio of 40\%, leaving every node and edge with potential gaps in their neighboring connections. To reconstruct these missing neighbors, each node and edge embedding must absorb and interpret its local context, meaning each embedding needs to understand the specific details of its immediate surroundings. Our approach of high-ratio random masking and subsequent reconstruction enables us to do away with the limitations posed by the size and shape of subgraphs used for prediction, and as a result, the node and edge embeddings derived from our task are encouraged to gain an understanding of local contextual details.
Secondly, with the removal of nodes or edges at high ratios, the remaining nodes and edges could be regarded as a set of subgraphs tasked with predicting the entire graph. This presents a more intricate graph-to-graph forecasting task compared to other self-directed pre-training tasks, which generally grasp global graph details using smaller graphs or motifs as their prediction targets. Our ``intense'' pre-training task of graph masking and rebuilding provides us with a broader perspective for learning superior node and edge embeddings capable of capturing the intricate details of houses at both the individual node/edge and holistic graph levels.

\noindent\textbf{Graph Transformer Encoder.} 
The encoder in our proposed system acts as a transformative bridge, converting the initial attributes of visible, unmasked nodes and edges into their corresponding embeddings within latent feature spaces. This process involves the node and edge aspects of the encoder which incorporate the proposed graph modeling block and a multi-head node attention mechanism. These features are designed in the vein of the Transformer architecture, which is a method renowned for its ability to effectively model sequential data. The Transformer-style design is founded on the operational procedure detailed in Figure \ref{fig:framework} and Figure \ref{fig:block}.
To elucidate, the graph modeling block of the encoder is devised to effectively extract the intrinsic characteristics of the graph structure, taking into account the interdependencies between nodes and edges. This block aids in the creation of robust representations that encapsulate the holistic relationship dynamics within the graph. Further, the multi-head node attention mechanism affords our system the capability to focus on diverse features in the node's local neighborhood, which can be critical for understanding the node's context and role within the overall graph structure.

In this way, our encoder is not merely a tool for translating initial attributes into latent feature spaces, but also an innovative engine that incorporates specialized components to learn, understand, and capture the nuanced dynamics of the graph. Through this process, the encoder creates potent embeddings that pave the way for efficient and effective downstream task processing.

\noindent\textbf{Graph Transformer Decoder.} 
Our decoding process begins with accepting an ensemble of reorganized room representations. This ensemble is comprised of two key components: (i) the embeddings of unmasked nodes and edges generated by the encoder, and (ii) the mask tokens that symbolize nodes and edges which have been excluded or removed from the initial structure. In our unique approach, the decoder not only fuses these two components but also restores their initial sequence in the source input graphs by integrating corresponding positional embeddings.
While the architecture of the decoder mirrors the Transformer-style of the encoder, it is designed with more compactness in mind. Specifically, it assembles $M$ ($M{=}2$ in our experiments) graph modeling blocks, where $M$ is significantly less than $L$, the number utilized by the encoder. The reason for this design choice lies in its application: during the pre-training phase, the decoder is tasked with house reconstruction, while only the encoder is engaged to generate house representations for subsequent prediction tasks.

This design insight is derived from the MAE approach, suggesting that the model's performance is not adversely impacted by a decoder that is either narrower (uses fewer nodes and edges) or shallower (uses fewer layers). As a result, in our asymmetric design of the encoder-decoder structure, the entire graph's nodes and edges are processed only at the stage of the lightweight decoder. This strategic decision considerably reduces the model's computational and memory requirements during the pre-training phase, ensuring efficient and sustainable processing.

\noindent \textbf{Optimization Objective.}
Both branches of our model - the node and edge branches - are tasked with reconstructing house graphs. They do this by predicting the attributes of the nodes and edges that have been masked or hidden. To achieve this, we append a linear layer to the end of each decoder's output. The output dimensions of this linear layer are set to match the total feature size of either the node (for the node branch) or the edge (for the edge branch).
This reconstruction process for both nodes and edges is a complex task, involving high-dimensional multi-label predictions. One of the advantages of this approach is that it can mitigate the issue of ambiguity in our predictions, by providing a more precise and focused predictive framework. Drawing inspiration from MAE, we calculate the pre-training loss based on the masked tokens. This approach leads us to define our final pre-training loss as follows:
\begin{equation}
\begin{aligned}
    \mathcal{L}_{pre-training} &= \mathcal{L}_{node} + \mathcal{L}_{edge} \\
    &= \sum_{n \in N_{mask}} \mathcal{L}_{ce}(N_n, Y_n) \\
    & + \sum_{(i, j)\in E_{mask}}  \mathcal{L}_{ce}(E_{(i, j)}, Y_{(i, j)}),
    \end{aligned}
\end{equation}
where $\mathcal{L}_{node}$ and $\mathcal{L}_{edge}$ denote the loss functions for the node and edge branches, respectively. $N_{mask}$ and $E_{mask}$ symbolize the collection of masked nodes and edges, respectively. The cross-entropy loss, $\mathcal{L}_{ce}$, is derived from the contrast between the predicted node features, $N_n$, and their corresponding ground truths, $Y_n$, or between the predicted edge features, $E_{(i, j)}$, and their corresponding ground truths, $Y_{(i, j)}$.

\begin{figure}[!t] \small
	\centering
	\includegraphics[width=1\linewidth]{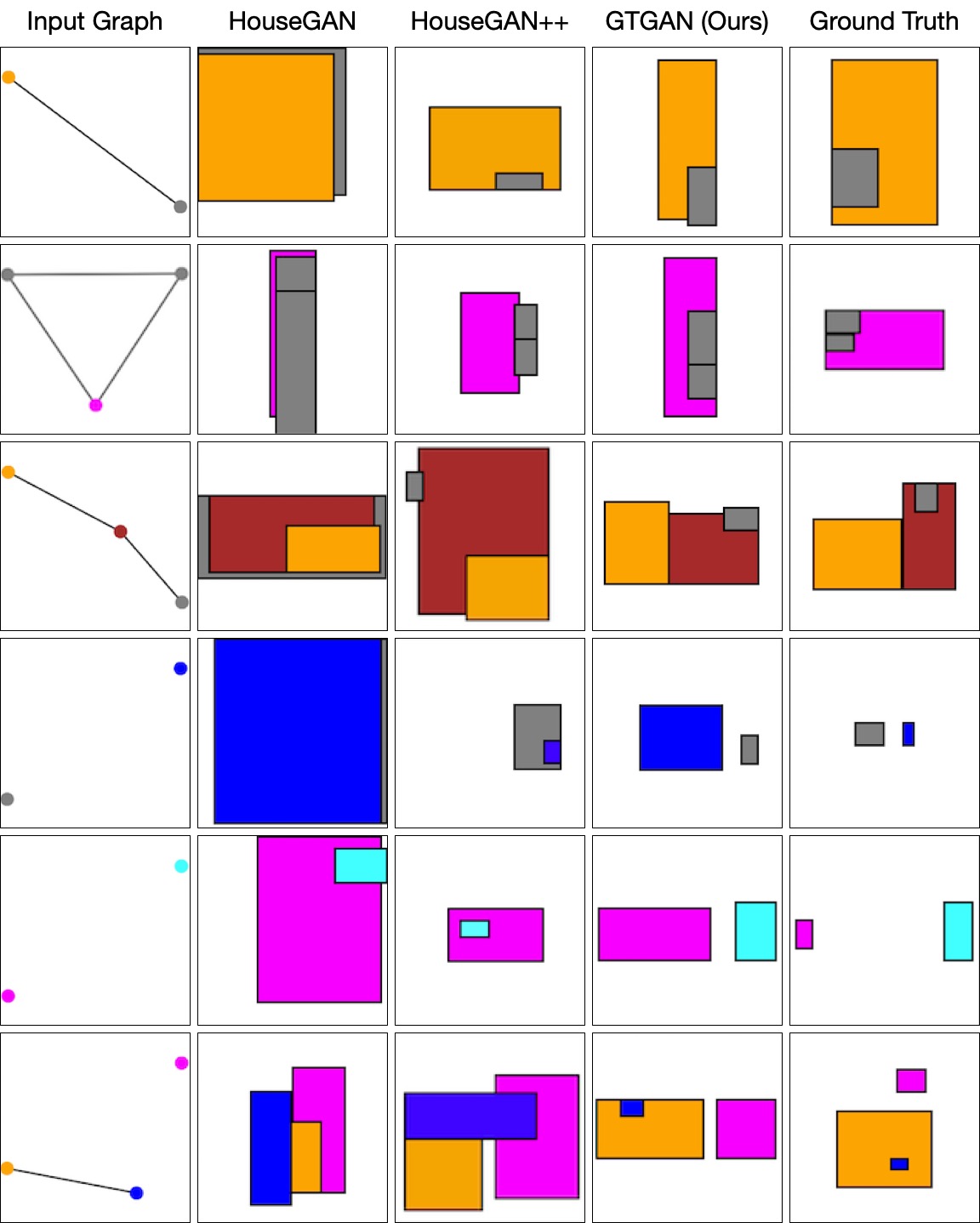}
	\caption{Visualization results compared with HouseGAN~\cite{nauata2020house} and HouseGAN++ \cite{nauata2021house} on ``1-3'' subset. The last three rows contain non-connected nodes.}
	\label{fig:results_A}
	\vspace{-0.4cm}
\end{figure}

\subsection{Fine-Tuning Strategy}
In the prior section \ref{sec:3}, we've extensively explained our fine-tuning strategy's methodology. For downstream tasks, our approach shifts to employ solely the graph Transformer encoder (GTE), marking a departure from the use of both encoder and decoder during the pre-training phase.
This change in strategy is fundamentally driven by the nature of the data feeding into the model. During the pre-training phase, the model is presented with incomplete graphs due to the masking process. However, for downstream tasks, the scenario changes drastically. The graphs fed into the model are fully complete, devoid of any masking.
This difference in input data between the pre-training and fine-tuning stages underscores the adaptability of our model. By focusing on the graph Transformer encoder during the downstream tasks, we aim to harness its capacity for handling complete graphs, extracting and processing the rich and complex data they present.
\section{Experiments}

The proposed method can be applied to different graph-based generative tasks such as house layout generation \cite{nauata2020house}, house roof generation \cite{qian2021roof}, and \hao{building layout generation \cite{chang2021building}}. In this section, we present experimental results and analysis on these three tasks.

\begin{figure}[!t] \small
	\centering
	\includegraphics[width=1\linewidth]{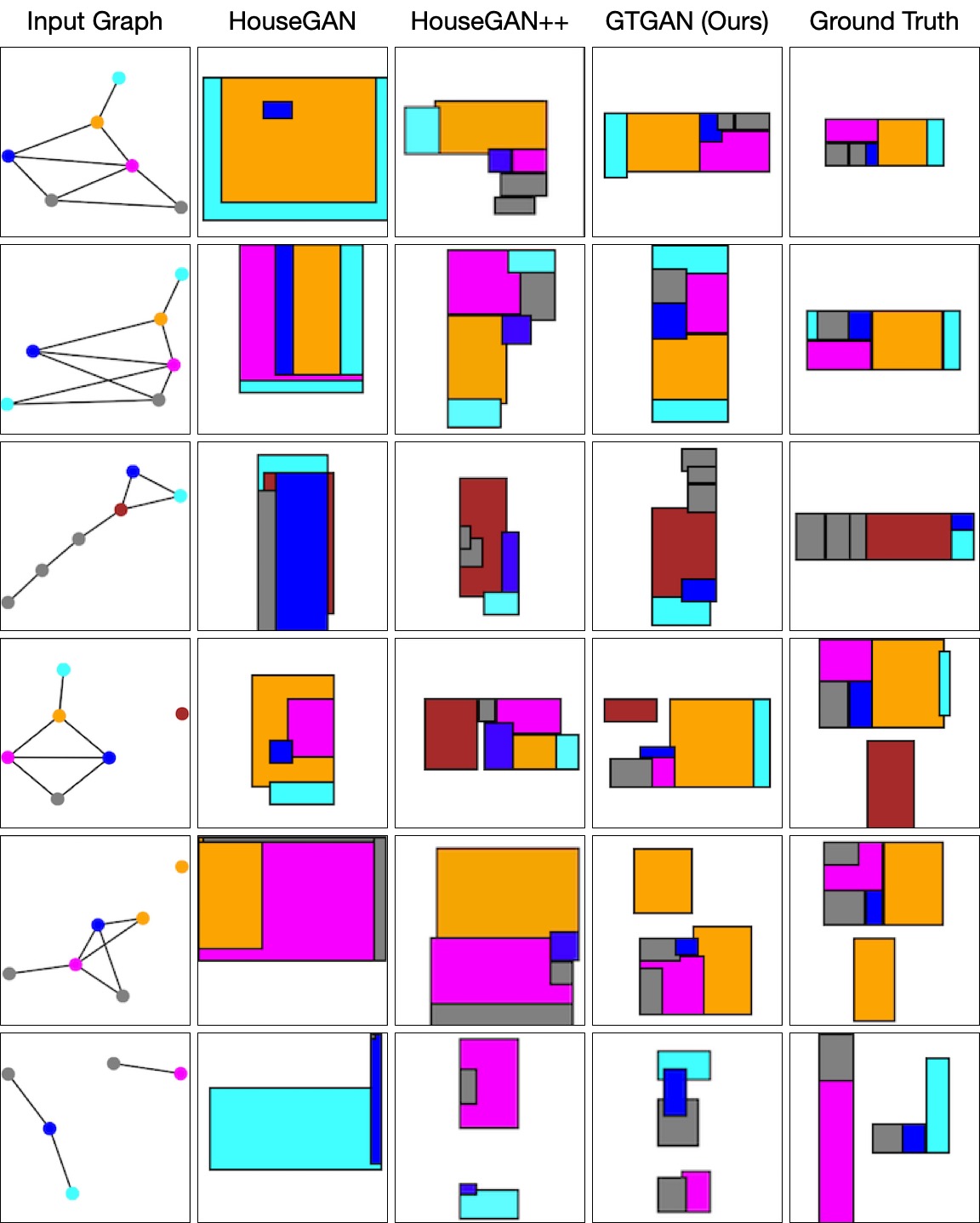}
	\caption{Visualization results compared with HouseGAN~\cite{nauata2020house} and HouseGAN++ \cite{nauata2021house} on  ``4-6'' subset. The last three rows contain non-connected nodes.}
	\label{fig:results_B}
	\vspace{-0.4cm}
\end{figure}

\begin{figure}[!t] \small
	\centering
	\includegraphics[width=1\linewidth]{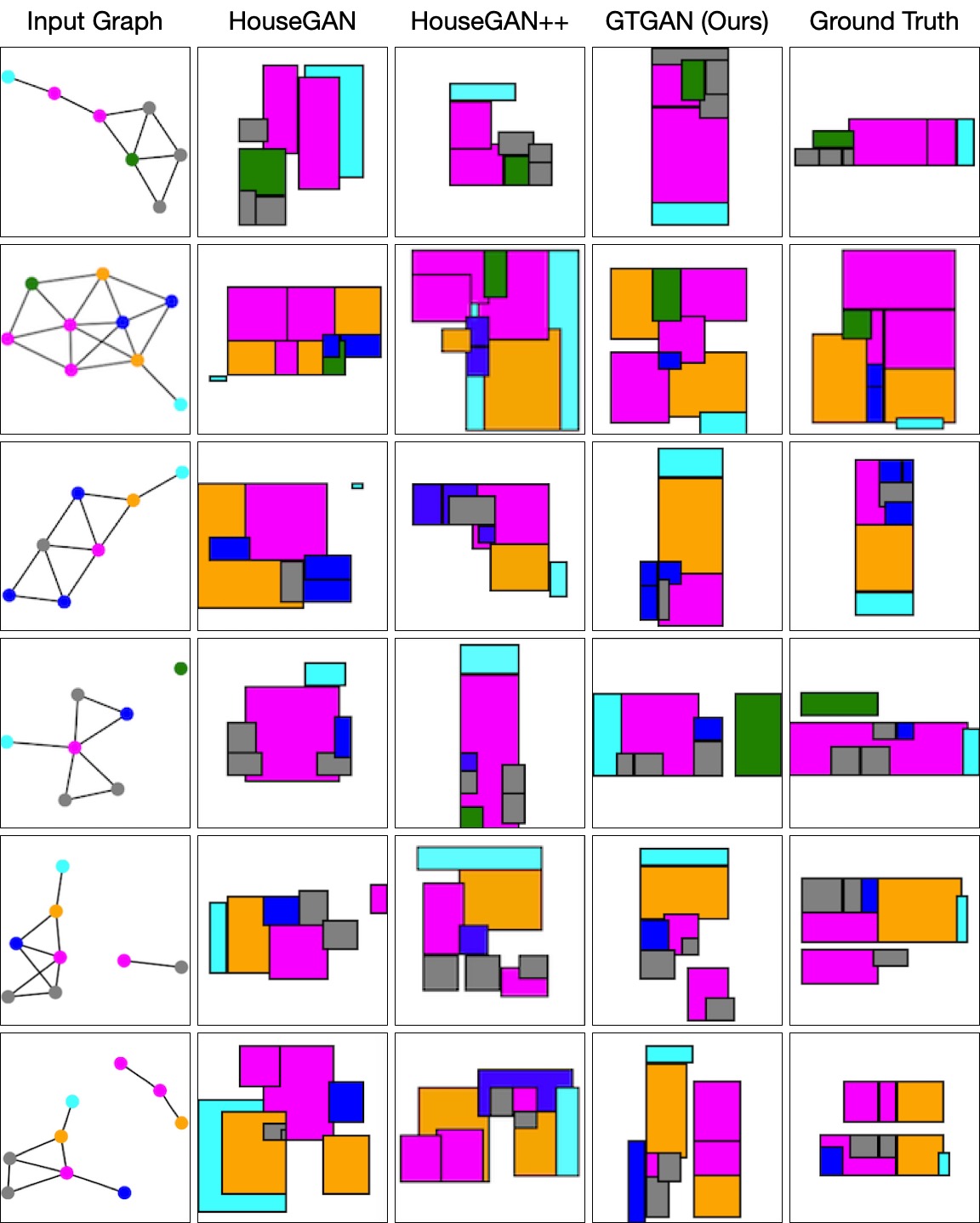}
	\caption{Visualization results compared with HouseGAN~\cite{nauata2020house} and HouseGAN++ \cite{nauata2021house} on  ``7-9'' subset. The last three rows contain non-connected nodes.}
	\label{fig:results_C}
	\vspace{-0.4cm}
\end{figure}

\begin{table*}[!ht] \small
	\centering
	\resizebox{1\linewidth}{!}{% 
		\begin{tabular}{rcccccccccccc} \toprule
			\multirow{2}{*}{Method}  & Realism $\uparrow$ & \multicolumn{5}{c}{Diversity $\downarrow$} & \multicolumn{5}{c}{Compatibility $\downarrow$} & \multirow{2}{*}{\hao{Training Time (h) $\downarrow$}} \\ \cmidrule(lr){2-2}  \cmidrule(lr){3-7} \cmidrule(lr){8-12} 
			& All Groups & 1-3 & 4-6   & 7-9 & 10-12  & 13+  & 1-3  & 4-6  & 7-9 & 10-12 & 13+      \\ \hline	
			CNN-only & -0.54 & 13.2 & 26.6 & 43.6 & 54.6 & 90.0 & 0.4 & 3.1 & 8.1 & 15.8 &34.7 & \hao{-}\\
			GCN         & 0.14 & 18.6 & 17.0 & 18.1 & 22.7 & 31.5 &  0.1 & 0.8 &  {\color{cyan} 2.3} & {\color{cyan} 3.2} & {\color{cyan} 3.7} & \hao{-}\\
			Ashual et al.~\cite{ashual2019specifying} & -0.55 & 64.0 & 92.2& 87.6& 122.8& 149.9&  0.2 & 2.7& 6.2& 19.2& 36.0 & \hao{-}\\
			Johnson et al.~\cite{johnson2018image}  & -0.58 & 69.8& 86.9& 80.1& 117.5& 123.2& 0.2 & 2.6& 5.2& 17.5& 29.3 & \hao{-}\\
			HouseGAN~\cite{nauata2020house}       & 0.17 & 13.6&  9.4 & 14.4& 11.6 & 20.1 & 0.1 &  1.1 & 2.9&  3.9 & 10.8 & {\color{blue}40}\\
			HouseGAN++$\ast$ \cite{nauata2021house} & {\color{cyan}0.19} & {\color{cyan} 11.8} & {\color{cyan} 7.6} & {\color{cyan} 12.2} & {\color{cyan} 10.1} & {\color{cyan} 18.3} & {\color{cyan} 0.08} & {\color{cyan} 0.77} &  2.52 & 3.65 & 7.43 & {\color{cyan}46}\\
			GTGAN \cite{tang2023graph} & {\color{blue} 0.25} & {\color{blue} 7.1}  & {\color{blue} 5.4}   & {\color{blue} 9.6}  & {\color{blue} 7.5} & {\color{blue} 16.9}  &  {\color{blue} 0.06} & {\color{blue} 0.62} & {\color{blue} 2.14}  & {\color{blue} 2.63}  & {\color{blue} 3.42} & 52 \\
   GTGAN++ (Ours) & {\color{orange} 0.29}& {\color{orange} 6.2} & {\color{orange} 4.0} & {\color{orange} 7.9}& {\color{orange} 5.3} & {\color{orange} 13.5} & {\color{orange} 0.04} & {\color{orange} 0.51} & {\color{orange} 1.89} & {\color{orange}  2.05} &  {\color{orange} 3.18} & {\color{orange}8+24}\\
			\bottomrule	
	\end{tabular}}
	\caption{Quantitative results and \hao{training time (h)} of house layout generation on the LIFULL HOME's dataset. The training time is tested on the same machine. The colors {\color{orange} orange}, {\color{blue} blue}, and {\color{cyan} cyan} represent the first, the second, and third best results, respectively. HouseGAN++$\ast$ \cite{nauata2021house} is reproduced by us.}
	\label{tab:house_reuslts}
	\vspace{-0.4cm}
\end{table*}

\begin{figure}[!t] \small
	\centering
	\includegraphics[width=1\linewidth]{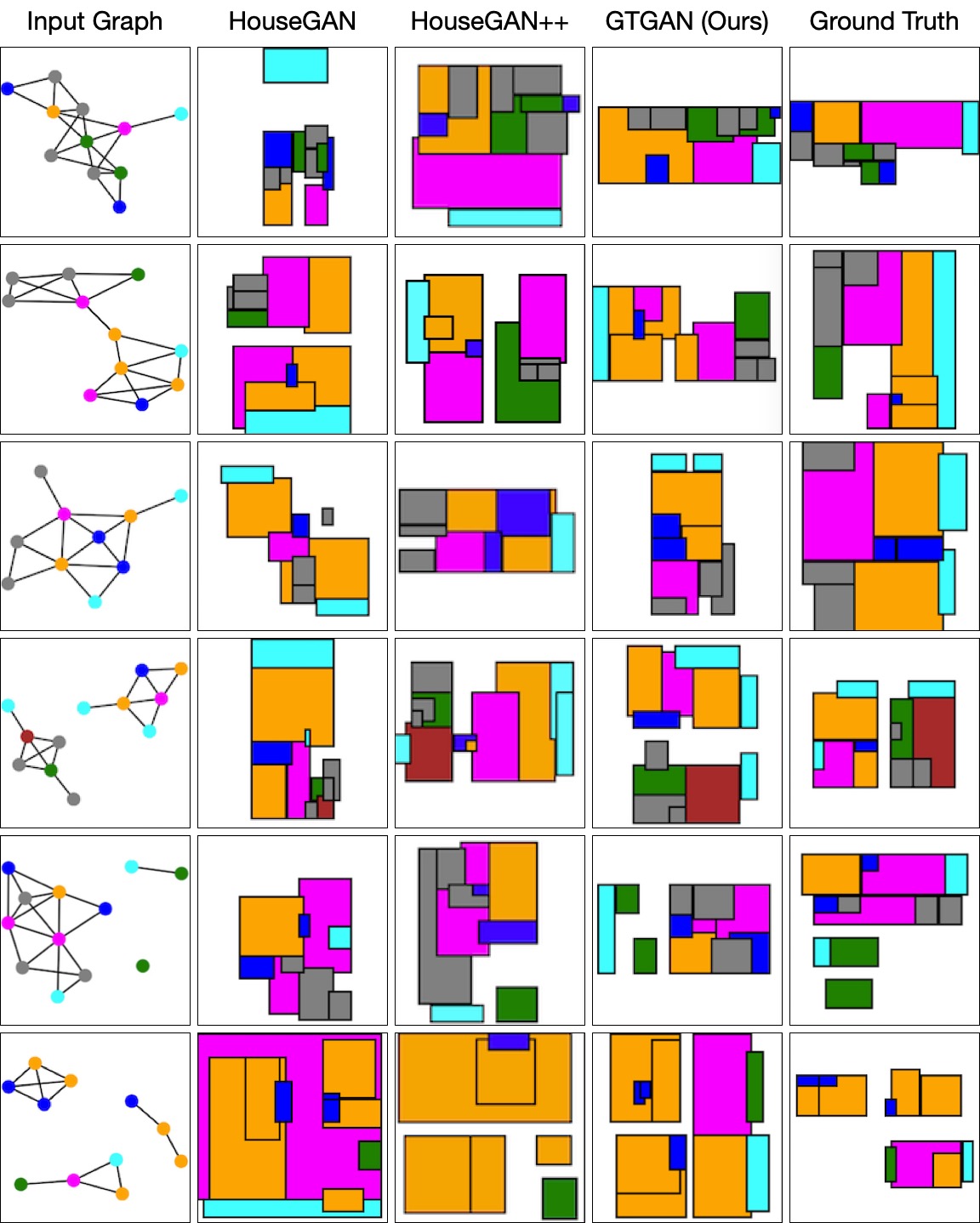}
	\caption{Visualization results compared with HouseGAN~\cite{nauata2020house} and HouseGAN++ \cite{nauata2021house} on ``10-12'' subset. The last three samples contain non-connected nodes.}
	\label{fig:results_D}
	\vspace{-0.4cm}
\end{figure}

\begin{figure}[!t] \small
	\centering
	\includegraphics[width=1\linewidth]{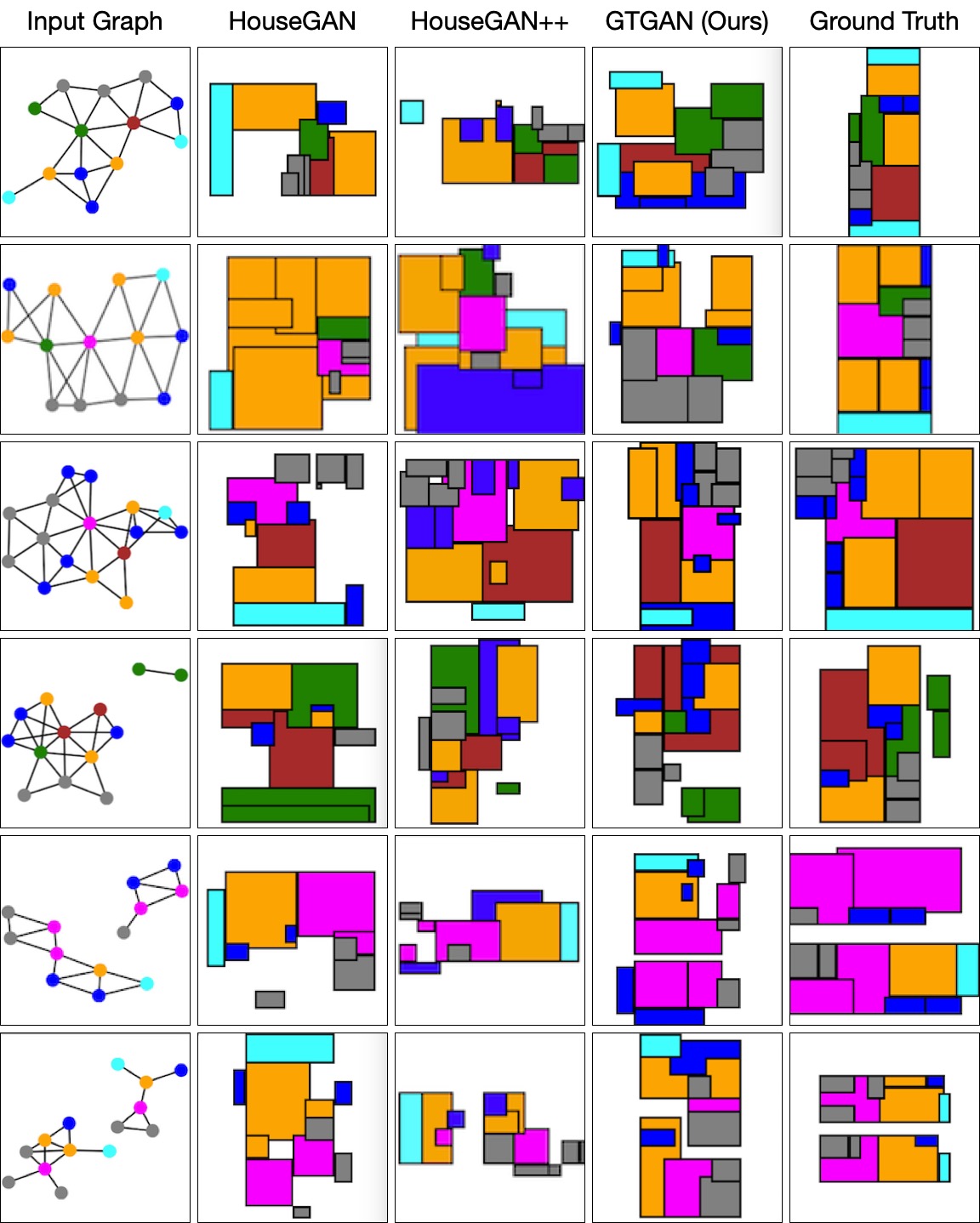}
	\caption{Visualization results compared with HouseGAN~\cite{nauata2020house} and HouseGAN++ \cite{nauata2021house} on ``13+'' subset. The last three samples contain non-connected nodes.}
	\label{fig:results_E}
	\vspace{-0.4cm}
\end{figure}

\subsection{Results on House Layout Generation}

\noindent \textbf{Datasets.}
We follow HouseGAN \cite{nauata2020house} and conduct house layout generation experiments on the LIFULL HOME's dataset, which have different rooms, i.e., living room, kitchen, bedroom, bathroom, closet, balcony, corridor, dining room, laundry room, and unknown.

\noindent \textbf{Evaluation Metrics.}
We follow HouseGAN~\cite{nauata2020house} and adopt realism, diversity, and compatibility metrics to evaluate the performance of the proposed method.
Specifically, we follow~\cite{nauata2020house} and divide the training samples into five subsets based on the number of rooms, i.e., 1-3, 4-6, 7-9, 10-12, and 13+. 
When generating layouts in a subset, we train the models while excluding samples in the same subset so that they cannot simply memorize. 
\begin{itemize}[leftmargin=*]
\item 1) We use the average user rating (12 Ph.D. students and 10 professional architects) to measure realism. 
	We provide 75 generated results with ground truths or 75 results generated by another method for comparison.
	A subject can give four ratings, i.e., better (+1), equally good (+1), worse (-1), or equally bad (-1).
 
\item 2) The Fr\'echet inception distance (FID) \cite{heusel2017gans} measures the diversity with the rasterized layout images.
	We rasterize a layout by setting the background to white and then sorting the rooms in decreasing order of area, finally painting each room with a color based on its room type. We follow \cite{nauata2020house} and use 5,000 samples to compute the FID metric.
 
\item 3) The compatibility of the bubble diagram is determined by the graph editing distance \cite{abu2015exact} between the input bubble diagram and the bubble diagram constructed from the generated layout.
\end{itemize}

\noindent \textbf{Quantitative Comparisons.}
To evaluate the effectiveness of the proposed method on house layout generation, we compare it with several leading methods, i.e., CNN-only, GCN, Ashual et al.~\cite{ashual2019specifying}, Johnson et al.~\cite{johnson2018image}, HouseGAN~\cite{nauata2020house}, and HouseGAN++ \cite{nauata2021house}.
We follow the same setups in \cite{nauata2020house} to reproduce the results of Ashual et al.~\cite{ashual2019specifying} and Johnson et al.~\cite{johnson2018image} for fair comparisons.
Table \ref{tab:house_reuslts} shows our main results on the five subsets. 
Note that we train HouseGAN++ \cite{nauata2021house} on this dataset using the public source code for a fair comparison, which is denoted as HouseGAN++$\ast$. 
We observe that the proposed methods outperform other competing methods in all the metrics, validating the effectiveness of our methods.

\noindent \textbf{Qualitative Comparisons.}
We compare the proposed GTGAN with the most related methods, i.e., HouseGAN~\cite{nauata2020house} and HouseGAN++ \cite{nauata2021house}.
The visualization results on the five subsets are shown in Figures~\ref{fig:results_A},~\ref{fig:results_B},~\ref{fig:results_C},~\ref{fig:results_D} and \ref{fig:results_E}. 
It is easy to tell that GTGAN generates more realistic and reasonable house layouts than the leading methods, i.e., HouseGAN and HouseGAN++.
For instance, HouseGAN generates improper room sizes or shapes for certain room types, e.g., the closet, the kitchen, and the closet in the last three rows of Figure~\ref{fig:results_A}(left), respectively, are too big. 
Moreover, HouseGAN generates misalignment of rooms, e.g., the balcony, the kitchen, and the living room in the first three samples of Figure~\ref{fig:results_A}(middle) do not align well with other rooms.
Lastly, both HouseGAN and  HouseGAN++ cannot generate non-connected rooms well, e.g., the corridor, the bathroom, and the bedroom in the last three samples of Figure~\ref{fig:results_A}(right) are not accurately generated because they are not connected to other rooms.
In contrast, the proposed method alleviates all three problems to a certain extent and generates more realistic and reasonable house layouts.

\begin{figure*}[!t] \small
	\centering
	\includegraphics[width=1\linewidth]{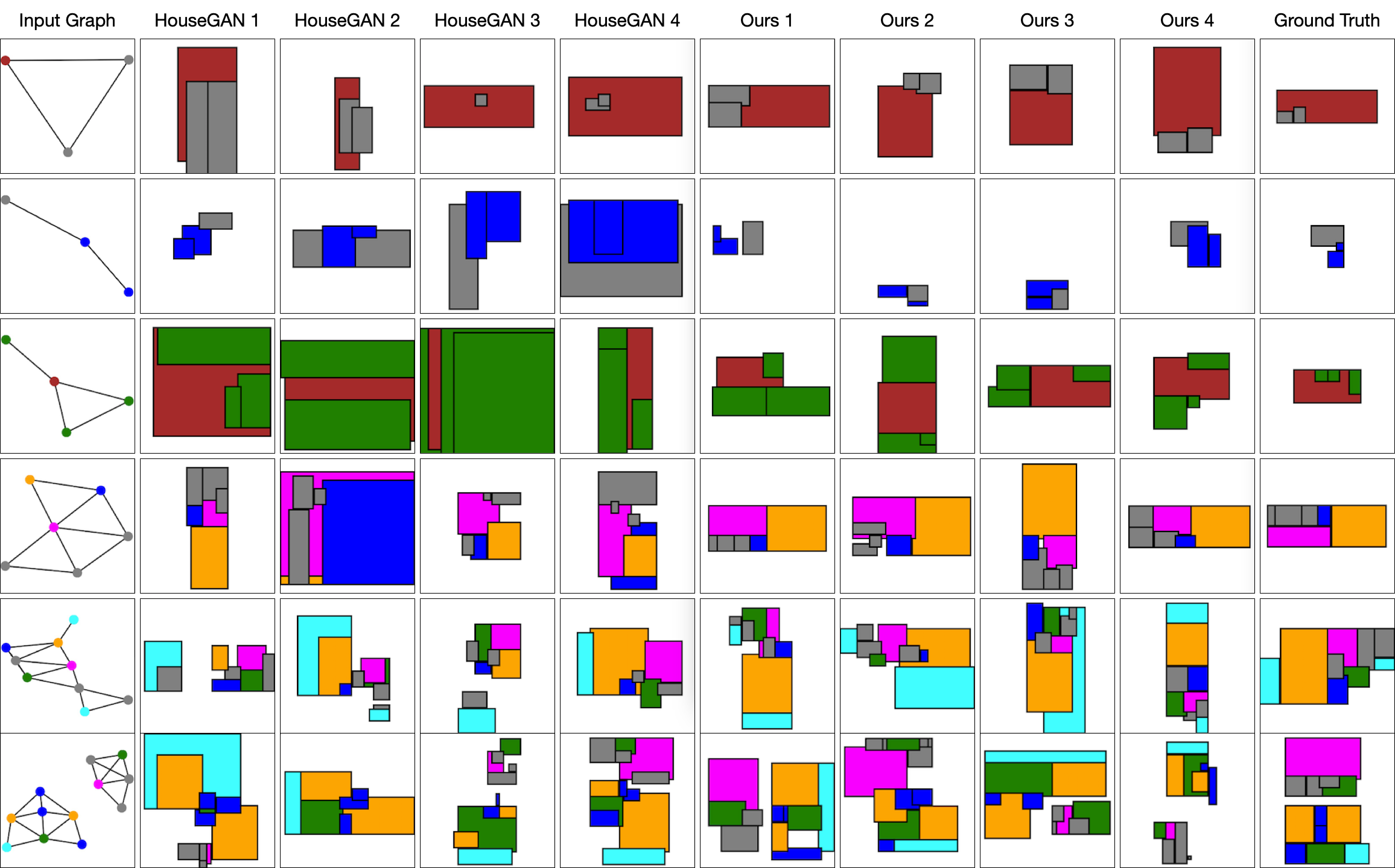}
	\caption{Diversity evaluation compared with HouseGAN~\cite{nauata2020house}. We generate four different house layout results using the same graph.}
	\label{fig:results_multi}
	\vspace{-0.4cm}
\end{figure*}

\noindent \textbf{Diversity Generation.}
For each input bubble diagram, we generate four different house layout variations. 
The results compared with the state-of-the-art HouseGAN~\cite{nauata2020house} are shown in Figure~\ref{fig:results_multi}.
We observe that HouseGAN is not capable of generating variations and tends to collapse into fewer modes.
In contrast, the proposed method has a better diversity generation with more reasonable house layouts.

\noindent \textbf{\hao{Training Time.}}
\hao{We also provide the training time (h) in Table \ref{tab:house_reuslts}. Because of the proposed graph masked modeling method, GTGAN++ takes less than half the training time of GTGAN to converge. Specifically, GTGAN training takes about 52 hours, while GTGAN++ only takes 24 hours to complete the training, plus the 8 hours of the pre-training phase, so the total time to train GTGAN++ is about 32 hours. This is far faster than both HouseGAN (40h) and HouseGAN++ (46h) methods.}

\begin{figure*}[!t] \small
	\centering
	\includegraphics[width=1\linewidth]{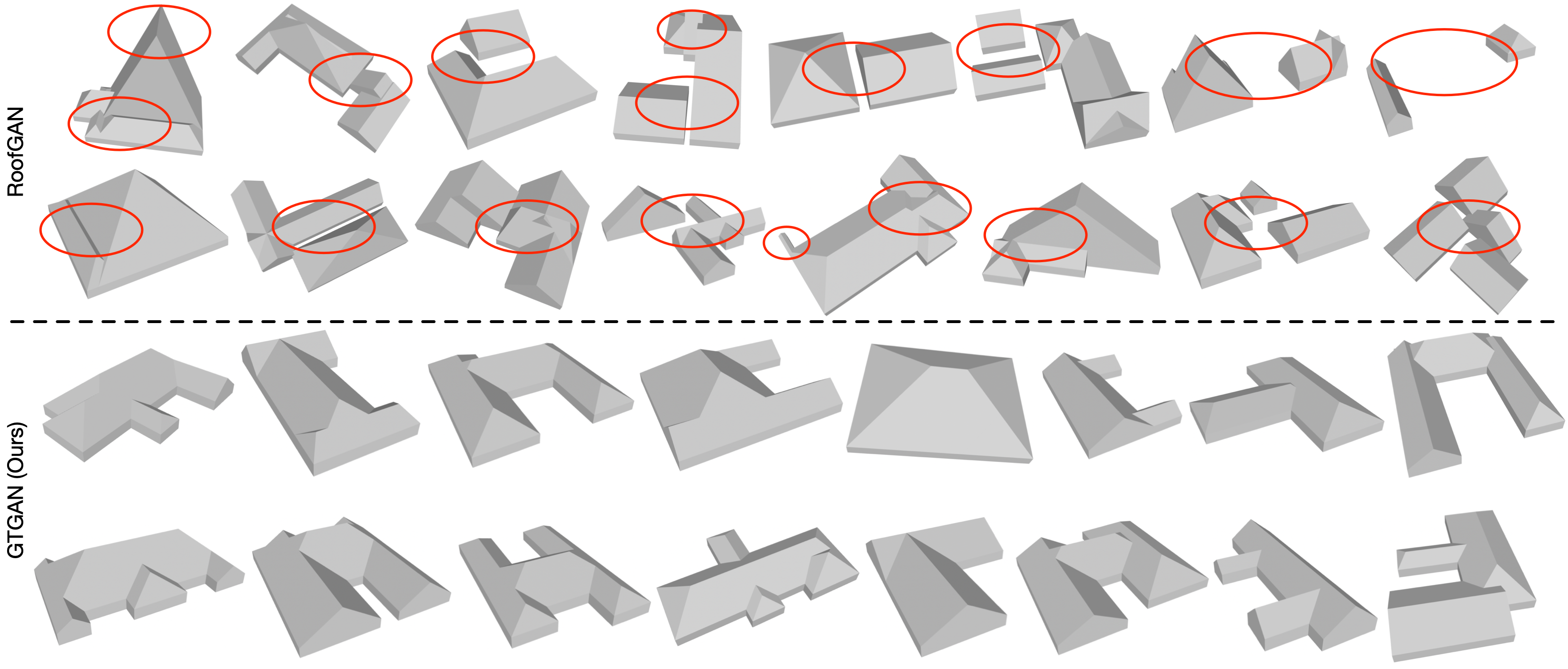}
	\caption{Visualization results compared with the proposed GTGAN (\textbf{bottom two rows}) and RoofGAN~(\textbf{top two rows}). We see that GTGAN can generate more realistic roof structures than RoofGAN. Red ovals highlight non-realistic roof structures generated by RoofGAN.}
	\label{fig:results_roof}
	\vspace{-0.4cm}
\end{figure*}

\subsection{Results on House Roof Generation}

\begin{figure}[!t] \small
	\centering
	\includegraphics[width=1\linewidth]{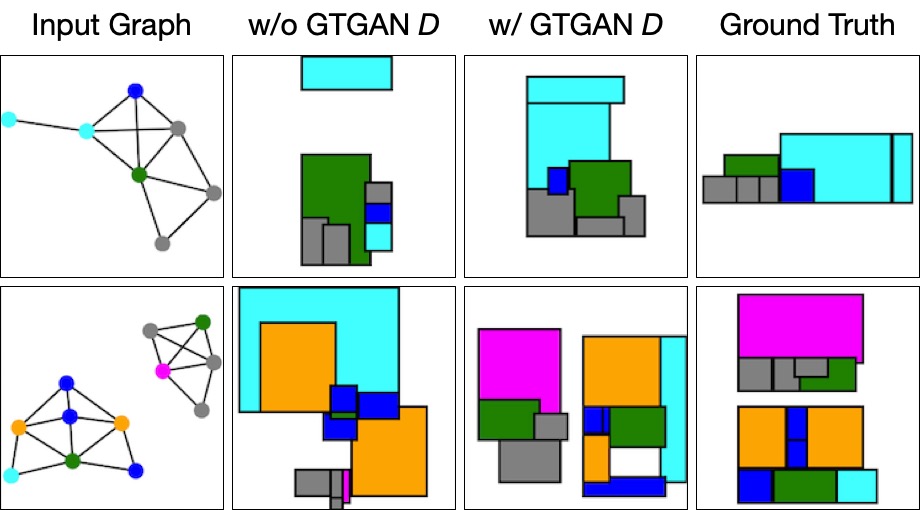}
	\caption{Comparison between w/o and w/ our GTGAN $D$. }
	\label{fig:node_discriminator}
	\vspace{-0.4cm}
\end{figure}

\noindent \textbf{Datasets and Evaluation Metrics.}
We follow RoofGAN \cite{qian2021roof} and conduct extensive experiments on the CAD-style roof geometry dataset proposed in \cite{qian2021roof}. We follow \cite{qian2021roof} and use the FID \cite{heusel2017gans} and the minimum matching distance (RMMD) as the evaluation metrics for a fair comparison.

\noindent \textbf{Quantitative Comparisons.}
To evaluate the effectiveness of GTGAN on house roof generation, we compare it with four leading methods, i.e., PQ-Net \cite{wu2020pq}, HouseGAN \cite{nauata2020house}, \hao{HouseGAN++ \cite{nauata2021house}}, and RoofGAN \cite{qian2021roof}.
Table \ref{tab:roof_reuslts} shows the comparison results on both 3 and 4 primitives.
When generating roofs, we follow \cite{qian2021roof} and split the training and test sets based on
the number of primitives to prevent simply copying and pasting.
We observe that the proposed methods outperform the other three competing methods in both metrics,  demonstrating the effectiveness of our methods.

\begin{table}[!t] \small
	\centering
	 	% \resizebox{0.8\linewidth}{!}{% 
	\begin{tabular}{rcccc} \toprule
		\multirow{2}{*}{Method} & \multicolumn{2}{c}{3 Primitives} & \multicolumn{2}{c}{4 Primitives} \\ \cmidrule(lr){2-3} \cmidrule(lr){4-5}
		& FID $\downarrow$ & RMMD $\downarrow$	& FID $\downarrow$ & RMMD $\downarrow$ \\ \midrule
		PQ-Net \cite{wu2020pq} & 13.0 & 10.4 & 14.6 & 12.9 \\
		HouseGAN \cite{nauata2020house} & 27.5 & 8.5 & 27.2 & 12.5 \\
  \hao{HouseGAN++ \cite{nauata2021house}} & \hao{25.2} & \hao{8.2} & \hao{24.3} & \hao{12.1} \\
		RoofGAN \cite{qian2021roof} & {\color{cyan}11.1} & {\color{cyan}7.5} & {\color{cyan}13.8} & {\color{cyan}10.9} \\ 
		GTGAN \cite{tang2023graph} & {\color{blue}9.3} & {\color{blue}5.5} & {\color{blue}9.6} & {\color{blue}7.2}\\ 
 GTGAN++ (Ours) & {\color{orange} 7.5} & {\color{orange} 4.2} & {\color{orange} 7.1} & {\color{orange} 5.6} \\ \bottomrule
	\end{tabular}
	\caption{Quantitative results of house roof generation on the CAD-style roof geometry dataset. The colors {\color{orange} orange}, {\color{blue} blue}, and {\color{cyan} cyan} represent the first, the second, and third best results, respectively.}
	\label{tab:roof_reuslts}
	\vspace{-0.2cm}
\end{table}

\begin{table}[!t] \small
	\centering
	 	% \resizebox{0.8\linewidth}{!}{% 
	\begin{tabular}{rcc} \toprule
		Method & FID $\downarrow$ & Connectivity Accuracy $\uparrow$ \\ \midrule
		HouseGAN \cite{nauata2020house} & 17.6003 & 0.403 \\
            HouseGAN++ \cite{nauata2021house} & 12.1845 & 0.494 \\
            BuildingGAN \cite{chang2021building} & {\color{cyan}0.0845} & {\color{cyan}0.569} \\
		GTGAN \cite{tang2023graph} & {\color{blue}0.0732} & {\color{blue}0.581}\\
            GTGAN++ (Ours) & {\color{orange}0.0528} & {\color{orange}0.617} \\ \bottomrule
	\end{tabular}
	\caption{\hao{Quantitative results of building layout generation. The colors {\color{orange} orange}, {\color{blue} blue}, and {\color{cyan} cyan} represent the first, the second, and third best results, respectively.}}
	\label{tab:building_reuslts}
	\vspace{-0.4cm}
\end{table}

\noindent \textbf{Qualitative Comparisons.}
We compare GTGAN with the most related method, i.e., RoofGAN \cite{nauata2020house}.
The visualization results are shown in Figure \ref{fig:results_roof}. 
Clearly, we observe that GTGAN generates more realistic and reasonable roof structures than the leading method RoofGAN.
For instance, RoofGAN generates isolated, too-long, too-high, or too-thin roofs. 
It also produces poor relationships between different components, which results in unrealistic polygonal shapes and topology.
In contrast, GTGAN alleviates all these problems to a certain extent and generates more complex and realistic combinations of roof primitives. 
Moreover, GTGAN also produces more diverse roofs, which is another advantage of our method.

\subsection{\hao{Results on Building Layout Generation}}

\noindent\textbf{\hao{Datasets and Evaluation Metrics.}}
\hao{Follong BuildingGAN \cite{chang2021building}, we use the same data set for experiments to make a fair comparison. This dataset contains 120,000 volumetric designs for commercial buildings. Moreover, we follow BuildingGAN and employ FID \cite{heusel2017gans} and connectivity accuracy as our evaluation metrics.}

\begin{table*}[!t] \small
	\centering
	% \resizebox{1\linewidth}{!}{% 
		\begin{tabular}{ccccccc} \toprule
			\# & Generator    & Discriminator  & FID $\downarrow$ & Compatibility  $\downarrow$ \\ \midrule	
			B1 & HouseGAN  \cite{nauata2020house}  & HouseGAN  \cite{nauata2020house} & 11.6  & 3.90 \\ 
			B2 & GTGAN & HouseGAN  \cite{nauata2020house} & 10.3 & 3.49 \\
			B3 & HouseGAN   \cite{nauata2020house} & GTGAN & 9.5  & 3.22 \\
			B4 & GTGAN & GTGAN & 8.2 &  2.95 \\  \hline
			B5 & GTGAN w/o NNA & GTGAN & 9.4 & 3.46 \\
			B6 & GTGAN w/o CNA & GTGAN & 9.7 & 3.68 \\
			B7 & GTGAN w/o GMB & GTGAN &  9.5 & 3.61 \\
			B8 & GTGAN w/ Transformer Layers & GTGAN & 8.9 & 3.45 \\ \hline
			B9 & GTGAN w/ Eq.~\eqref{eq:eq1} & GTGAN & 9.4 & 3.29 \\ 
			B10 & GTGAN w/ Eq.~\eqref{eq:eq2} & GTGAN & 9.8 & 3.32 \\ \hline
			B11  &  \multicolumn{2}{c}{B4 + Graph-Based Cycle-Consistency Loss}  & 7.5 &  2.63 \\ \hline
                B12  &  \multicolumn{2}{c}{B11 + Graph Node Masking}  & 5.9 & 2.36 \\ 
                B13  &  \multicolumn{2}{c}{B11 + Graph Edge Masking}  & 5.7 & 2.28 \\ 
                B14  &  \multicolumn{2}{c}{B11 + Graph Node and Edge Masking}  & \textbf{5.3} & \textbf{2.05} \\ \bottomrule
	\end{tabular}
	\caption{Ablation study of the proposed method on house layout generation.}
	\label{tab:abla}
	\vspace{-0.4cm}
\end{table*}

\noindent\textbf{\hao{State-of-the-Art Comparisons.}}
\hao{To evaluate the effectiveness of the proposed method on building layout generation, we compare it with three leading methods, i.e., HouseGAN \cite{nauata2020house}, \hao{HouseGAN++ \cite{nauata2021house}}, and BuildingGAN \cite{chang2021building}. 
To expand HouseGAN, HouseGAN++, GTGAN, and GTGAN++ into the realm of 3D, it involves appending the story index of each program node to its respective feature.
Results are shown in Table \ref{tab:building_reuslts}, we see that the proposed GTGAN++ achieves the best results on both metrics, validating the effectiveness of the proposed graph masked modeling method.}

\subsection{Ablation Study}
We conduct extensive ablation studies on house layout generation (``10-12'' subset) to evaluate the effectiveness of each component of the proposed method.

\noindent \textbf{Baselines Models.}
GTGAN has 14 baselines, as shown in Table~\ref{tab:abla}: 
(1) B1 is our baseline combining HouseGAN $G$ and HouseGAN $D$ (i.e., the original HouseGAN~\cite{nauata2020house}).
(2) B2 adopts the combination of GTGAN $G$ and HouseGAN $D$.
(3) B3 combines HouseGAN $G$ and GTGAN $D$. 
(4) B4 employs both GTGAN $G$ and GTGAN $D$.
(5) B5 is our baseline without using NNA.
(6) B6 is our baseline without using CNA.
(7) B7 is our baseline without using GMB.
(8) B8 is our baseline using Transformer layers instead of Conv-MPN layers.
(9) B9 is the variation using Eq.~\eqref{eq:eq1} instead of Eq.~\eqref{eq:eq}.
(10) B10 is the variation using Eq.~\eqref{eq:eq2}  instead of Eq.~\eqref{eq:eq}.
(11) B11 is using the proposed graph-based cycle-consistency loss upon B4.
(12) B12 uses the proposed graph node masking strategy upon B11.
(13) B13 employs the proposed graph edge masking strategy upon B11.
(14) B13 is our full model (i.e., GTGAN++), which employs both graph node and edge masking strategies upon B11.

\noindent \textbf{Ablation Analysis.}
The results of the ablation study, shown in Table~\ref{tab:abla}, prove that our graph Transformer generator $G$ and node classification-based discriminator $D$ improve the generation performance over the baseline models, validating the effectiveness of the proposed framework.
Specifically, when using our generator, B2 yields further improvements over B1, meaning that our generator learns local and global relations across connected and non-connected nodes more effectively, confirming our design motivation.
B3 outperforms B1, demonstrating the importance of using our discriminator to generate semantically consistent rooms according to the input graph nodes.
We show the comparison results in Figure \ref{fig:node_discriminator}.
We see that using our discriminator is helping to preserve the room information in the generated floorplans, leading to a better house layout.
Moreover, we observe that B4 generates significantly better results than B2 and B3, further confirming our network design.

We also observe that not using NNA or CNA significantly reduces performance in B5 or B6, which validates the effectiveness of both NNA and CNA.
Also, without using the proposed GMB in B7, the performance drops a lot on both metrics.
Meanwhile, using Transformer layers (B8) other than Conv-MPN layers slightly reduces the performance, which means that using a mixed model of CNN and Transformer can achieve better results.
When we use Eq.~\eqref{eq:eq1} and Eq.~\eqref{eq:eq2} instead of Eq.~\eqref{eq:eq} in B9 and B10, the performance drops slightly, which also proves the rationality of our model design.
Also, B11 significantly outperforms B4, clearly demonstrating the effectiveness of the proposed graph-based cycle-consistency loss.

\begin{figure}[!t] \small
	\centering
	\includegraphics[width=1\linewidth]{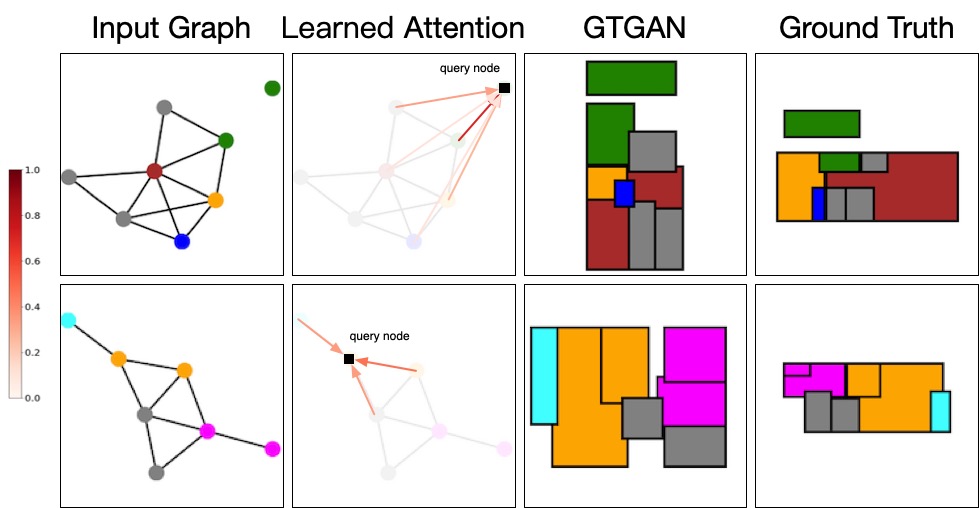}
	\caption{Visualization of the learned node attention.}
	\label{fig:node_attention}
	\vspace{-0.4cm}
\end{figure}

Lastly, we see that B14 outperforms the other two baselines (i.e., B12 and B13) in terms of both metrics, suggesting the efficacy of our designed approach which combines both node and edge masking strategies. Furthermore, even with the implementation of only one masking strategy (i.e., B12 or B13), the model can still greatly surpass B11 in performance. This underlines the potency of our graph masking in absorbing and interpreting information from the house layout graphs.

\noindent \textbf{Visulization of Attention Weights.} We show two examples of the learned node attention in Figure~\ref{fig:node_attention}. The query nodes are colored in black, whereas the edges are colored according to the magnitude of the attention weights, which can be referred to by the color bar on the left. We can observe that the proposed method indeed learns the relationships between nodes.

\noindent \textbf{Influence of Hyperparameter.}
We first set $\lambda_2{=}0.1$ and then investigate the influence of $\lambda_1$ in Eq.~\eqref{eq:loss}  on the performance of our full model. 
As shown in Table \ref{tab:abla_para}, when $\lambda_1{=}1$, our model achieves the best generation performance. 
Moreover, we set $\lambda_1{=}1$, and then investigate the influence of $\lambda_2$ on the performance of our full model. 
As shown in Table \ref{tab:abla_para2}, when $\lambda_2{=}0.1$, our model achieves the best generation performance. 
Therefore, we adopt $\lambda_1{=}1$ and $\lambda_2{=}0.1$ in all experiments.

\begin{figure}[!t] \small
	\centering
	\includegraphics[width=0.8\linewidth]{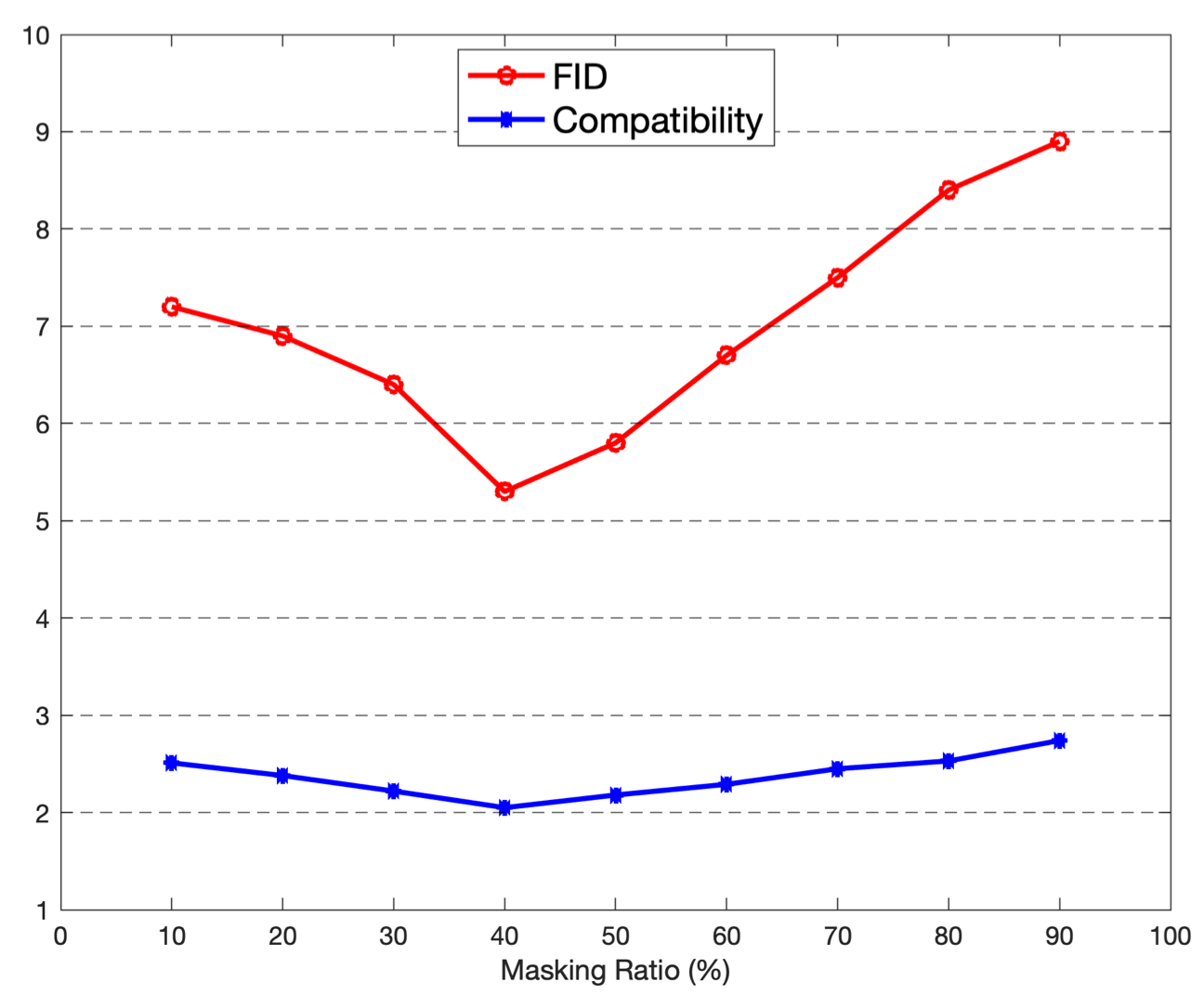}
	\caption{Ablation study of the masking ratio.}
	\label{fig:ratio}
	\vspace{-0.2cm}
\end{figure}

\begin{table}[!t] \small
	\centering
	\resizebox{0.8\linewidth}{!}{% 
		\begin{tabular}{ccccccc} \toprule
			$\lambda_1$         & 0.01    & 0.1  & 0.2   & 0.5       & 1   & 10 \\ \midrule	
			FID $\downarrow$ &  9.8    & 11.4 & 17.7&  10.5     & \textbf{7.5}& 13.4 \\  \bottomrule
	\end{tabular}}
	\caption{Ablation study of the $\lambda_1$ in Eq.~\eqref{eq:loss}.}
	\label{tab:abla_para}
	\vspace{-0.2cm}
\end{table}

\begin{table}[!t] \small
	\centering
	\resizebox{0.8\linewidth}{!}{% 
		\begin{tabular}{ccccccc} \toprule
			$\lambda_2$         & 0.01    & 0.1  & 0.2   & 0.5       & 1   & 10 \\ \midrule	
			FID $\downarrow$ &   9.4        & \textbf{7.5} &  8.3 & 9.8 & 10.4 & 11.5 \\  \bottomrule
	\end{tabular}}
	\caption{Ablation study of the $\lambda_2$ in Eq.~\eqref{eq:loss}.}
	\label{tab:abla_para2}
	\vspace{-0.4cm}
\end{table}

\noindent \textbf{Influence of Masking Ratio.}
In an effort to assess the impact of varying masking ratios, we conduct pre-training of our proposed method using a spectrum of masking ratios from 10\% to 90\% specifically in the context of house layout generation. Our results in Figure \ref{fig:ratio} show that a masking ratio of 40\% renders the most favorable performance. This indicates that a relatively high masking ratio for nodes and edges during pre-training provides a more extensive learning opportunity for the remaining nodes and edges to assimilate intricate information in their embeddings. However, we note that when the masking ratio surpassed 40\%, the residual nodes and edges lacked sufficient information to rebuild the entire graph, which started to detrimentally affect the quality of the learned embeddings.

\begin{figure}[!t] \small
	\centering
	\includegraphics[width=1\linewidth]{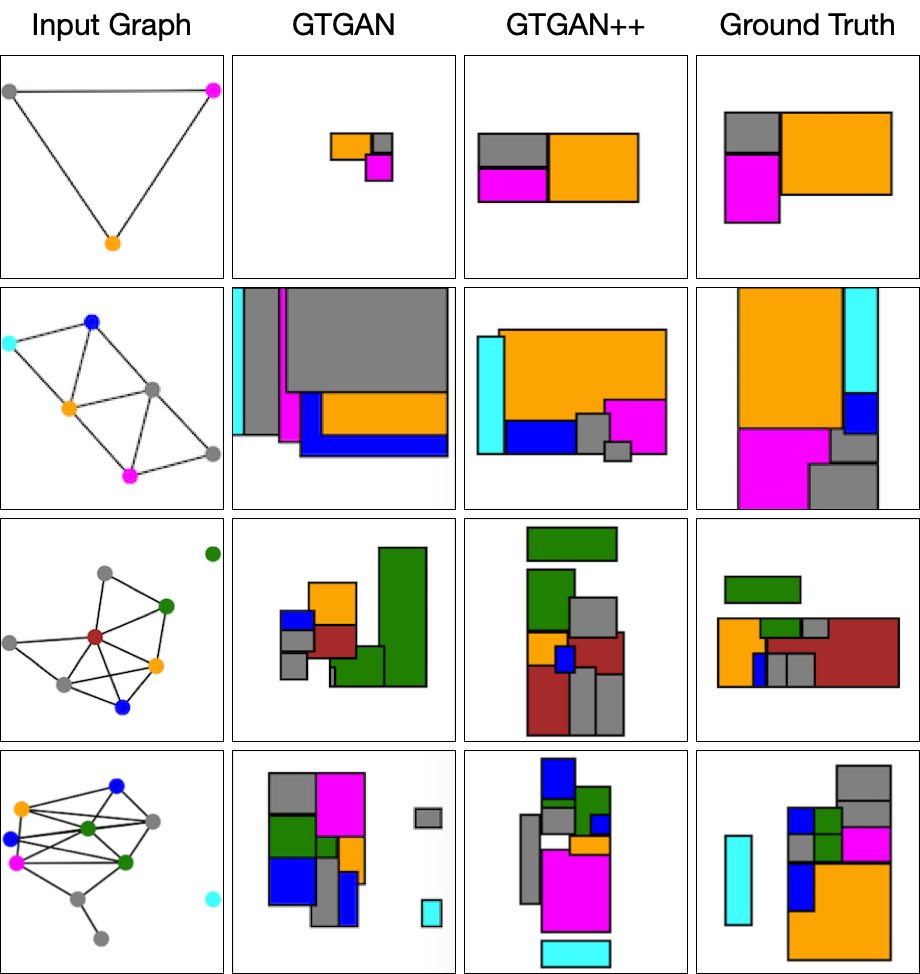}
 	\includegraphics[width=1\linewidth]{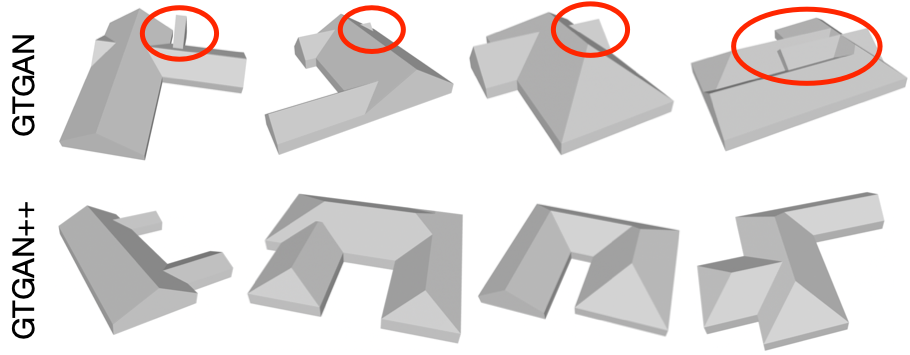}
	\caption{Comparison results of GTGAN and GTGAN++ on house layout generation (\textbf{top four rows}) and roof generation (\textbf{bottom two rows}). }
	\label{fig:gtgan}
	\vspace{-0.4cm}
\end{figure}

\noindent \textbf{GTGAN vs. GTGAN++.}
From Tables \ref{tab:house_reuslts} and \ref{tab:roof_reuslts}, we can see that GTGAN++ achieves much better results than GTGAN on all metrics in both house layout generation and roof generation tasks, meaning that our proposed graph masked modeling strategy is effective.
Moreover, we note that GTGAN++ generates much better results than GTGAN on both tasks, as shown in Figure \ref{fig:gtgan}. This also demonstrates the effectiveness of our graph masked modeling strategy.

\section{Limitations and Conclusion}
\noindent \hao{\textbf{Limitations.} The datasets used in existing papers do not capture the distribution of real architectural layouts. This makes our method sensitive to the viewpoint and style of the graph representation. It is possible to address this limitation by collecting a real dataset of architectural rooms and roofs and associated 3D building meshes.}

\noindent \textbf{Conclusion.}
With this work, we are the first to explore using a Transformer-based architecture for graph-constrained \hao{architectural layout generation} tasks. 
We provide four contributions, i.e., a graph Transformer-based generator, a node classification-based discriminator, a graph-based cycle-consistency loss, and a graph masked modeling strategy.
The first is employed to model local and global relations across connected and non-connected nodes in a graph.
The second component is used to preserve high-level, semantically discriminative features for different house components.
The third is used to preserve relative spatial relationships between ground truth and generated graphs.
The last one aims to ensure the model is well-equipped to handle downstream tasks efficiently, providing a robust foundation during the pre-training stage and flexible adaptation during the fine-tuning stage.
Extensive experiments in terms of both human and automatic evaluation demonstrate that the proposed method achieves remarkably better performance than existing approaches on three challenging generation tasks.  
We believe that our framework makes an essential step towards automated architectural layout design.\\

\noindent\textbf{Acknowledgments.} This work was partly supported by the ETH Zurich General Fund (OK), the MUR PNRR project FAIR (PE00000013) funded by the
NextGenerationEU, the PRIN project CREATIVE (Prot. 2020ZSL9F9), and by the EU Horizon project ELIAS (No. 101120237).

%\clearpage
\footnotesize
\bibliographystyle{IEEEtran}
\bibliography{egbib}

\begin{thebibliography}{10}
\providecommand{\url}[1]{#1}
\csname url@samestyle\endcsname
\providecommand{\newblock}{\relax}
\providecommand{\bibinfo}[2]{#2}
\providecommand{\BIBentrySTDinterwordspacing}{\spaceskip=0pt\relax}
\providecommand{\BIBentryALTinterwordstretchfactor}{4}
\providecommand{\BIBentryALTinterwordspacing}{\spaceskip=\fontdimen2\font plus
\BIBentryALTinterwordstretchfactor\fontdimen3\font minus
  \fontdimen4\font\relax}
\providecommand{\BIBforeignlanguage}[2]{{%
\expandafter\ifx\csname l@#1\endcsname\relax
\typeout{** WARNING: IEEEtran.bst: No hyphenation pattern has been}%
\typeout{** loaded for the language `#1'. Using the pattern for}%
\typeout{** the default language instead.}%
\else
\language=\csname l@#1\endcsname
\fi
#2}}
\providecommand{\BIBdecl}{\relax}
\BIBdecl

\bibitem{wang2019planit}
K.~Wang, Y.-A. Lin, B.~Weissmann, M.~Savva, A.~X. Chang, and D.~Ritchie,
  ``Planit: Planning and instantiating indoor scenes with relation graph and
  spatial prior networks,'' \emph{ACM TOG}, vol.~38, no.~4, pp. 1--15, 2019.

\bibitem{hu2020graph2plan}
R.~Hu, Z.~Huang, Y.~Tang, O.~Van~Kaick, H.~Zhang, and H.~Huang, ``Graph2plan:
  Learning floorplan generation from layout graphs,'' \emph{ACM TOG}, vol.~39,
  no.~4, pp. 118--1, 2020.

\bibitem{ashual2019specifying}
O.~Ashual and L.~Wolf, ``Specifying object attributes and relations in
  interactive scene generation,'' in \emph{ICCV}, 2019.

\bibitem{johnson2018image}
J.~Johnson, A.~Gupta, and L.~Fei-Fei, ``Image generation from scene graphs,''
  in \emph{CVPR}, 2018.

\bibitem{nauata2020house}
N.~Nauata, K.-H. Chang, C.-Y. Cheng, G.~Mori, and Y.~Furukawa, ``House-gan:
  Relational generative adversarial networks for graph-constrained house layout
  generation,'' in \emph{ECCV}, 2020.

\bibitem{wu2020pq}
R.~Wu, Y.~Zhuang, K.~Xu, H.~Zhang, and B.~Chen, ``Pq-net: A generative part
  seq2seq network for 3d shapes,'' in \emph{CVPR}, 2020.

\bibitem{qian2021roof}
Y.~Qian, H.~Zhang, and Y.~Furukawa, ``Roof-gan: learning to generate roof
  geometry and relations for residential houses,'' in \emph{CVPR}, 2021.

\bibitem{vaswani2017attention}
A.~Vaswani, N.~Shazeer, N.~Parmar, J.~Uszkoreit, L.~Jones, A.~N. Gomez,
  L.~Kaiser, and I.~Polosukhin, ``Attention is all you need,'' in
  \emph{NeurIPS}, 2017.

\bibitem{dosovitskiy2020image}
A.~Dosovitskiy, L.~Beyer, A.~Kolesnikov, D.~Weissenborn, X.~Zhai,
  T.~Unterthiner, M.~Dehghani, M.~Minderer, G.~Heigold, S.~Gelly \emph{et~al.},
  ``An image is worth 16x16 words: Transformers for image recognition at
  scale,'' in \emph{ICLR}, 2021.

\bibitem{zheng2020rethinking}
S.~Zheng, J.~Lu, H.~Zhao, X.~Zhu, Z.~Luo, Y.~Wang, Y.~Fu, J.~Feng, T.~Xiang,
  P.~H. Torr \emph{et~al.}, ``Rethinking semantic segmentation from a
  sequence-to-sequence perspective with transformers,'' in \emph{CVPR}, 2021.

\bibitem{wang2020max}
H.~Wang, Y.~Zhu, H.~Adam, A.~Yuille, and L.-C. Chen, ``Max-deeplab: End-to-end
  panoptic segmentation with mask transformers,'' in \emph{CVPR}, 2021.

\bibitem{wang2021transbts}
W.~Wang, C.~Chen, M.~Ding, H.~Yu, S.~Zha, and J.~Li, ``Transbts: Multimodal
  brain tumor segmentation using transformer,'' in \emph{MICCAI}, 2021.

\bibitem{carion2020end}
N.~Carion, F.~Massa, G.~Synnaeve, N.~Usunier, A.~Kirillov, and S.~Zagoruyko,
  ``End-to-end object detection with transformers,'' in \emph{ECCV}, 2020.

\bibitem{zhu2020deformable}
X.~Zhu, W.~Su, L.~Lu, B.~Li, X.~Wang, and J.~Dai, ``Deformable detr: Deformable
  transformers for end-to-end object detection,'' in \emph{ICLR}, 2021.

\bibitem{huang2020hand}
L.~Huang, J.~Tan, J.~Liu, and J.~Yuan, ``Hand-transformer: Non-autoregressive
  structured modeling for 3d hand pose estimation,'' in \emph{ECCV}, 2020.

\bibitem{huang2020hot}
L.~Huang, J.~Tan, J.~Meng, J.~Liu, and J.~Yuan, ``Hot-net: Non-autoregressive
  transformer for 3d hand-object pose estimation,'' in \emph{ACM MM}, 2020.

\bibitem{lin2020end}
K.~Lin, L.~Wang, and Z.~Liu, ``End-to-end human pose and mesh reconstruction
  with transformers,'' in \emph{CVPR}, 2021.

\bibitem{chen2020topological}
K.~Chen, J.~K. Chen, J.~Chuang, M.~V{\'a}zquez, and S.~Savarese, ``Topological
  planning with transformers for vision-and-language navigation,'' in
  \emph{CVPR}, 2021.

\bibitem{chang2021building}
K.-H. Chang, C.-Y. Cheng, J.~Luo, S.~Murata, M.~Nourbakhsh, and Y.~Tsuji,
  ``Building-gan: Graph-conditioned architectural volumetric design
  generation,'' in \emph{ICCV}, 2021.

\bibitem{tang2023graph}
H.~Tang, Z.~Zhang, H.~Shi, B.~Li, L.~Shao, N.~Sebe, R.~Timofte, and
  L.~Van~Gool, ``Graph transformer gans for graph-constrained house
  generation,'' in \emph{CVPR}, 2023.

\bibitem{goodfellow2014generative}
I.~Goodfellow, J.~Pouget-Abadie, M.~Mirza, B.~Xu, D.~Warde-Farley, S.~Ozair,
  A.~Courville, and Y.~Bengio, ``Generative adversarial nets,'' in
  \emph{NeurIPS}, 2014.

\bibitem{karras2018style}
T.~Karras, S.~Laine, and T.~Aila, ``A style-based generator architecture for
  generative adversarial networks,'' in \emph{CVPR}, 2019.

\bibitem{shaham2019singan}
T.~R. Shaham, T.~Dekel, and T.~Michaeli, ``Singan: Learning a generative model
  from a single natural image,'' in \emph{ICCV}, 2019.

\bibitem{tang2022local}
H.~Tang, L.~Shao, P.~H. Torr, and N.~Sebe, ``Local and global gans with
  semantic-aware upsampling for image generation,'' \emph{IEEE TPAMI}, vol.~45,
  no.~1, pp. 768--784, 2022.

\bibitem{tang2022multi}
H.~Tang, P.~H. Torr, and N.~Sebe, ``Multi-channel attention selection gans for
  guided image-to-image translation,'' \emph{IEEE TPAMI}, no.~01, pp. 1--16,
  2022.

\bibitem{mirza2014conditional}
M.~Mirza and S.~Osindero, ``Conditional generative adversarial nets,''
  \emph{arXiv preprint arXiv:1411.1784}, 2014.

\bibitem{choi2017stargan}
Y.~Choi, M.~Choi, M.~Kim, J.-W. Ha, S.~Kim, and J.~Choo, ``Stargan: Unified
  generative adversarial networks for multi-domain image-to-image
  translation,'' in \emph{CVPR}, 2018.

\bibitem{han2017stackgan}
H.~Zhang, T.~Xu, H.~Li, S.~Zhang, X.~Wang, X.~Huang, and D.~Metaxas,
  ``Stackgan: Text to photo-realistic image synthesis with stacked generative
  adversarial networks,'' in \emph{ICCV}, 2017.

\bibitem{tao2023galip}
M.~Tao, B.-K. Bao, H.~Tang, and C.~Xu, ``Galip: Generative adversarial clips
  for text-to-image synthesis,'' in \emph{CVPR}, 2023.

\bibitem{tao2022df}
M.~Tao, H.~Tang, F.~Wu, X.-Y. Jing, B.-K. Bao, and C.~Xu, ``Df-gan: A simple
  and effective baseline for text-to-image synthesis,'' in \emph{CVPR}, 2022.

\bibitem{xu2022predict}
Z.~Xu, T.~Lin, H.~Tang, F.~Li, D.~He, N.~Sebe, R.~Timofte, L.~Van~Gool, and
  E.~Ding, ``Predict, prevent, and evaluate: Disentangled text-driven image
  manipulation empowered by pre-trained vision-language model,'' in
  \emph{CVPR}, 2022.

\bibitem{reed2016learning}
S.~E. Reed, Z.~Akata, S.~Mohan, S.~Tenka, B.~Schiele, and H.~Lee, ``Learning
  what and where to draw,'' in \emph{NeurIPS}, 2016.

\bibitem{tang2020xinggan}
H.~Tang, S.~Bai, L.~Zhang, P.~H. Torr, and N.~Sebe, ``Xinggan for person image
  generation,'' in \emph{ECCV}, 2020.

\bibitem{tang2022bipartite}
H.~Tang, L.~Shao, P.~H. Torr, and N.~Sebe, ``Bipartite graph reasoning gans for
  person pose and facial image synthesis,'' \emph{Springer IJCV}, pp. 1--15,
  2022.

\bibitem{tang2021total}
H.~Tang and N.~Sebe, ``Total generate: Cycle in cycle generative adversarial
  networks for generating human faces, hands, bodies, and natural scenes,''
  \emph{IEEE TMM}, vol.~24, pp. 2963--2974, 2021.

\bibitem{tang2020bipartite}
H.~Tang, S.~Bai, P.~H. Torr, and N.~Sebe, ``Bipartite graph reasoning gans for
  person image generation,'' in \emph{BMVC}, 2020.

\bibitem{tang2019multi}
H.~Tang, D.~Xu, N.~Sebe, Y.~Wang, J.~J. Corso, and Y.~Yan, ``Multi-channel
  attention selection gan with cascaded semantic guidance for cross-view image
  translation,'' in \emph{CVPR}, 2019.

\bibitem{park2019semantic}
T.~Park, M.-Y. Liu, T.-C. Wang, and J.-Y. Zhu, ``Semantic image synthesis with
  spatially-adaptive normalization,'' in \emph{CVPR}, 2019.

\bibitem{tang2020local}
H.~Tang, D.~Xu, Y.~Yan, P.~H. Torr, and N.~Sebe, ``Local class-specific and
  global image-level generative adversarial networks for semantic-guided scene
  generation,'' in \emph{CVPR}, 2020.

\bibitem{tang2021layout}
H.~Tang and N.~Sebe, ``Layout-to-image translation with double pooling
  generative adversarial networks,'' \emph{IEEE TIP}, vol.~30, pp. 7903--7913,
  2021.

\bibitem{tang2020dual}
H.~Tang, S.~Bai, and N.~Sebe, ``Dual attention gans for semantic image
  synthesis,'' in \emph{ACM MM}, 2020.

\bibitem{tang2023edge}
H.~Tang, X.~Qi, G.~Sun, D.~Xu, N.~Sebe, R.~Timofte, and L.~Van~Gool, ``Edge
  guided gans with contrastive learning for semantic image synthesis,'' in
  \emph{ICLR}, 2023.

\bibitem{tang2023edge2}
H.~Tang, G.~Sun, N.~Sebe, and L.~Van~Gool, ``Edge guided gans with multi-scale
  contrastive learning for semantic image synthesis,'' \emph{IEEE TPAMI}, 2023.

\bibitem{mejjati2018unsupervised}
Y.~A. Mejjati, C.~Richardt, J.~Tompkin, D.~Cosker, and K.~I. Kim,
  ``Unsupervised attention-guided image-to-image translation,'' in
  \emph{NeurIPS}, 2018.

\bibitem{tang2021attentiongan}
H.~Tang, H.~Liu, D.~Xu, P.~H. Torr, and N.~Sebe, ``Attentiongan: Unpaired
  image-to-image translation using attention-guided generative adversarial
  networks,'' \emph{IEEE TNNLS}, 2021.

\bibitem{dhamo2021graph}
H.~Dhamo, F.~Manhardt, N.~Navab, and F.~Tombari, ``Graph-to-3d: End-to-end
  generation and manipulation of 3d scenes using scene graphs,'' in
  \emph{ICCV}, 2021.

\bibitem{luo2020end}
A.~Luo, Z.~Zhang, J.~Wu, and J.~B. Tenenbaum, ``End-to-end optimization of
  scene layout,'' in \emph{CVPR}, 2020.

\bibitem{lee2020neural}
H.-Y. Lee, L.~Jiang, I.~Essa, P.~B. Le, H.~Gong, M.-H. Yang, and W.~Yang,
  ``Neural design network: Graphic layout generation with constraints,'' in
  \emph{ECCV}, 2020.

\bibitem{yamaguchi2021canvasvae}
K.~Yamaguchi, ``Canvasvae: Learning to generate vector graphic documents,'' in
  \emph{ICCV}, 2021.

\bibitem{arroyo2021variational}
D.~M. Arroyo, J.~Postels, and F.~Tombari, ``Variational transformer networks
  for layout generation,'' in \emph{CVPR}, 2021.

\bibitem{dai2022ao2}
L.~Dai, H.~Liu, H.~Tang, Z.~Wu, and P.~Song, ``Ao2-detr: Arbitrary-oriented
  object detection transformer,'' \emph{IEEE TCSVT}, 2022.

\bibitem{dong2023hotbev}
P.~Dong, Z.~Kong, X.~Meng, P.~Yu, Y.~Gong, G.~Yuan, H.~Tang, and Y.~Wang,
  ``Hotbev: Hardware-oriented transformer-based multi-view 3d detector for bev
  perception,'' in \emph{NeurIPS}, 2023.

\bibitem{dong2023speeddetr}
P.~Dong, Z.~Kong, X.~Meng, P.~Zhang, H.~Tang, Y.~Wang, and C.-H. Chou,
  ``Speeddetr: Speed-aware transformers for end-to-end object detection,'' in
  \emph{ICML}, 2023.

\bibitem{yang2021transformer}
G.~Yang, H.~Tang, M.~Ding, N.~Sebe, and E.~Ricci, ``Transformer-based attention
  networks for continuous pixel-wise prediction,'' in \emph{ICCV}, 2021.

\bibitem{li2022mhformer}
W.~Li, H.~Liu, H.~Tang, P.~Wang, and L.~Van~Gool, ``Mhformer: Multi-hypothesis
  transformer for 3d human pose estimation,'' in \emph{CVPR}, 2022.

\bibitem{li2023multi}
W.~Li, H.~Liu, H.~Tang, and P.~Wang, ``Multi-hypothesis representation learning
  for transformer-based 3d human pose estimation,'' \emph{Elsevier PR}, vol.
  141, p. 109631, 2023.

\bibitem{zeng2020learning}
Y.~Zeng, J.~Fu, and H.~Chao, ``Learning joint spatial-temporal transformations
  for video inpainting,'' in \emph{ECCV}, 2020.

\bibitem{neimark2021video}
D.~Neimark, O.~Bar, M.~Zohar, and D.~Asselmann, ``Video transformer network,''
  in \emph{ICCV}, 2021.

\bibitem{chopin2023interaction}
B.~Chopin, H.~Tang, N.~Otberdout, M.~Daoudi, and N.~Sebe, ``Interaction
  transformer for human reaction generation,'' \emph{IEEE TMM}, 2023.

\bibitem{chen2022geometry}
H.~Chen, H.~Tang, Z.~Yu, N.~Sebe, and G.~Zhao, ``Geometry-contrastive
  transformer for generalized 3d pose transfer,'' in \emph{AAAI}, 2022.

\bibitem{chen2021aniformer}
H.~Chen, H.~Tang, N.~Sebe, and G.~Zhao, ``Aniformer: Data-driven 3d animation
  with transformer,'' in \emph{BMVC}, 2021.

\bibitem{chen2023lart}
H.~Chen, H.~Tang, R.~Timofte, L.~Van~Gool, and G.~Zhao, ``Lart: Neural
  correspondence learning with latent regularization transformer for 3d motion
  transfer,'' in \emph{NeurIPS}, 2023.

\bibitem{kenton2019bert}
J.~D. M.-W.~C. Kenton and L.~K. Toutanova, ``Bert: Pre-training of deep
  bidirectional transformers for language understanding,'' in \emph{NAACL-HLT},
  2019.

\bibitem{he2022masked}
K.~He, X.~Chen, S.~Xie, Y.~Li, P.~Doll{\'a}r, and R.~Girshick, ``Masked
  autoencoders are scalable vision learners,'' in \emph{CVPR}, 2022.

\bibitem{xie2022simmim}
Z.~Xie, Z.~Zhang, Y.~Cao, Y.~Lin, J.~Bao, Z.~Yao, Q.~Dai, and H.~Hu, ``Simmim:
  A simple framework for masked image modeling,'' in \emph{CVPR}, 2022.

\bibitem{mahmood2021masked}
O.~Mahmood, E.~Mansimov, R.~Bonneau, and K.~Cho, ``Masked graph modeling for
  molecule generation,'' \emph{Nature communications}, vol.~12, no.~1, p. 3156,
  2021.

\bibitem{hou2022graphmae}
Z.~Hou, X.~Liu, Y.~Cen, Y.~Dong, H.~Yang, C.~Wang, and J.~Tang, ``Graphmae:
  Self-supervised masked graph autoencoders,'' in \emph{KDD}, 2022.

\bibitem{tian2023heterogeneous}
Y.~Tian, K.~Dong, C.~Zhang, C.~Zhang, and N.~V. Chawla, ``Heterogeneous graph
  masked autoencoders,'' in \emph{AAAI}, 2023.

\bibitem{zhang2020conv}
F.~Zhang, N.~Nauata, and Y.~Furukawa, ``Conv-mpn: Convolutional message passing
  neural network for structured outdoor architecture reconstruction,'' in
  \emph{CVPR}, 2020.

\bibitem{zhang2019self}
H.~Zhang, I.~Goodfellow, D.~Metaxas, and A.~Odena, ``Self-attention generative
  adversarial networks,'' in \emph{ICML}, 2019.

\bibitem{hendrycks2016gaussian}
D.~Hendrycks and K.~Gimpel, ``Gaussian error linear units (gelus),''
  \emph{arXiv preprint arXiv:1606.08415}, 2016.

\bibitem{kolotouros2019convolutional}
N.~Kolotouros, G.~Pavlakos, and K.~Daniilidis, ``Convolutional mesh regression
  for single-image human shape reconstruction,'' in \emph{CVPR}, 2019.

\bibitem{yu2019layout}
W.~Yu, X.~Liang, K.~Gong, C.~Jiang, N.~Xiao, and L.~Lin, ``Layout-graph
  reasoning for fashion landmark detection,'' in \emph{CVPR}, 2019.

\bibitem{kingma2014adam}
D.~P. Kingma and J.~Ba, ``Adam: A method for stochastic optimization,'' in
  \emph{ICLR}, 2015.

\bibitem{wang2024batmannet}
Z.~Wang, Z.~Feng, Y.~Li, B.~Li, Y.~Wang, C.~Sha, M.~He, and X.~Li, ``Batmannet:
  bi-branch masked graph transformer autoencoder for molecular
  representation,'' \emph{Briefings in Bioinformatics}, 2024.

\bibitem{nauata2021house}
N.~Nauata, S.~Hosseini, K.-H. Chang, H.~Chu, C.-Y. Cheng, and Y.~Furukawa,
  ``House-gan++: Generative adversarial layout refinement network towards
  intelligent computational agent for professional architects,'' in
  \emph{CVPR}, 2021.

\bibitem{heusel2017gans}
M.~Heusel, H.~Ramsauer, T.~Unterthiner, B.~Nessler, and S.~Hochreiter, ``Gans
  trained by a two time-scale update rule converge to a local nash
  equilibrium,'' in \emph{NeurIPS}, 2017.

\bibitem{abu2015exact}
Z.~Abu-Aisheh, R.~Raveaux, J.-Y. Ramel, and P.~Martineau, ``An exact graph edit
  distance algorithm for solving pattern recognition problems,'' in
  \emph{ICPRAM}, 2015.

\end{thebibliography}

\begin{IEEEbiography}[{\includegraphics[width=1in,height=1.25in,clip,keepaspectratio]{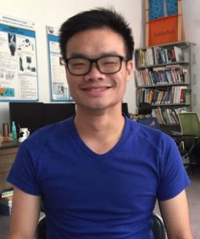}}]{Hao Tang} is currently a Postdoctoral with Computer Vision Lab, ETH Zurich, Switzerland.
He received the master’s degree from the School of Electronics and Computer Engineering, Peking University, China and the Ph.D. degree from Multimedia and Human Understanding Group, University of Trento, Italy.
He was a visiting scholar in the Department of Engineering Science at the University of Oxford. His research interests are deep learning, machine learning, and their applications to computer vision.
\end{IEEEbiography}

% \begin{IEEEbiography}[{\includegraphics[width=1in,height=1.25in,clip,keepaspectratio]{Figures/zzy.png}}]{Zhenyu Zhang} is a staff research scientist at Tencent Youtu Lab, where he works on computer vision, graphics and machine learning. He got the Ph.D degree from Department of Computer Science and Engineering, Nanjing University of Science and Technology in 2020, supervised by Jian Yang. In 2019, He spent 10 months as a visiting student at MHUG group in Unviversity of Trento, Italy, supervised by Nicu Sebe. He is interested in 3D reconstruction, neural rendering and implicit neural representation. 
% % Much of his research is about inferring the 3D model (depth, normal, mesh, point cloud, etc) and intrinsic cues (light, albedo, specular, roughness, etc) from images, and rendering high-realistic photo based on such information.
% \end{IEEEbiography}

\begin{IEEEbiography}[{\includegraphics[width=1in,height=1.25in,clip,keepaspectratio]{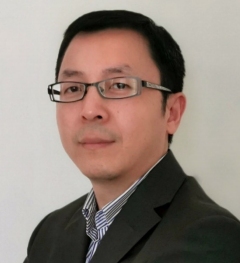}}]{Ling Shao} is a Distinguished Professor with the UCAS-Terminus AI Lab, University of Chinese Academy of Sciences, Beijing, China. He was the founding CEO and Chief Scientist of the Inception Institute of Artificial Intelligence, Abu Dhabi, UAE. He was also the Initiator, founding Provost and EVP of MBZUAI, UAE. His research interests include generative AI, vision and language, and AI for healthcare. He is a fellow of the IEEE, the IAPR, the BCS and the IET.
% His research interests include Computer Vision, Deep Learning/Machine Learning, Multimedia, and Image/Video Processing. 
	
\end{IEEEbiography}

\begin{IEEEbiography}[{\includegraphics[width=1in,height=1.25in,clip,keepaspectratio]{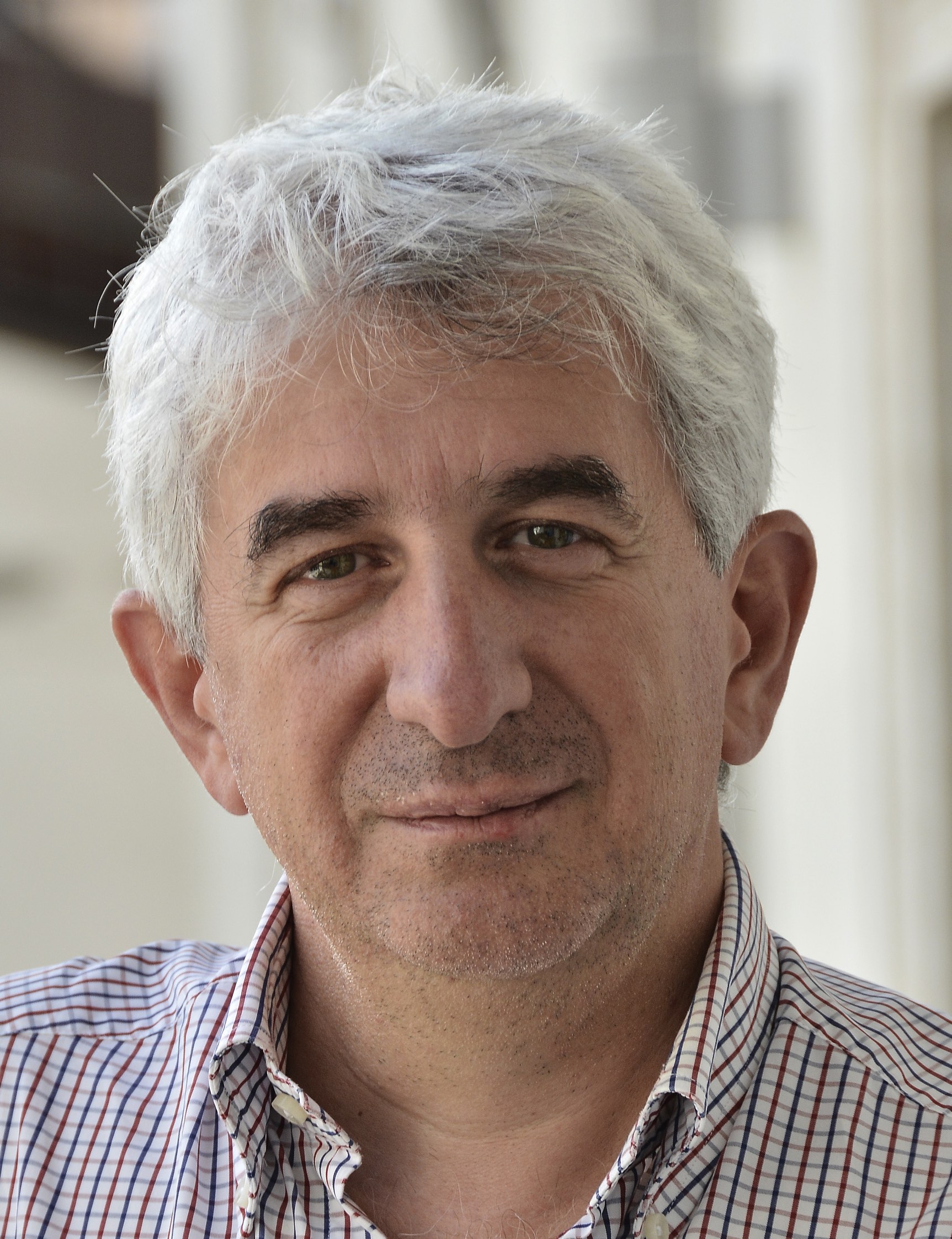}}]{Nicu Sebe} is Professor with the University of Trento, Italy, leading the research in the areas of multimedia information retrieval and human behavior understanding. He was the General Co-Chair of ACM Multimedia 2013 and 2022, and the Program Chair of ACM Multimedia 2007 and 2011, ECCV 2016, ICCV 2017 and ICPR 2020. He is a fellow of the International Association for Pattern Recognition (IAPr) and of the European Laboratory for Learning and Intelligent Systems (ELLIS).
\end{IEEEbiography}

\begin{IEEEbiography}[{\includegraphics[width=1in,height=1.25in,clip,keepaspectratio]{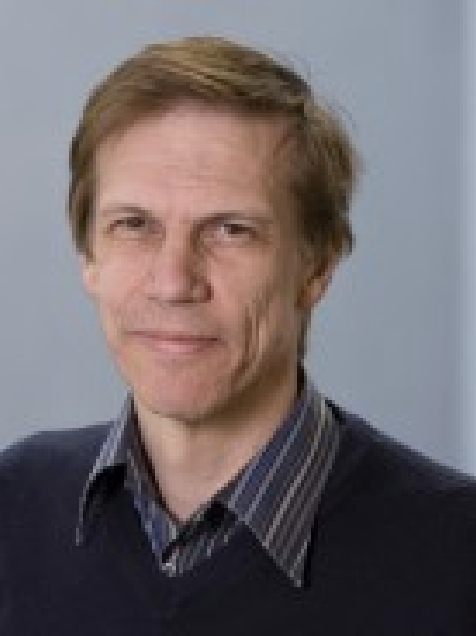}}]{Luc Van Gool} 
	received the degree in electromechanical engineering at the Katholieke Universiteit
	Leuven, in 1981. Currently, he is a professor at
	the Katholieke Universiteit Leuven in Belgium,
	the ETH Zurich in Switzerland, and the Sofia Un. St. Kliment Ohridski in Bulgaria. He leads computer vision research at all three places. 
	% He has been a program committee member of several major computer vision conferences. 
	His main interests include 3D reconstruction and modeling, object recognition, tracking,
	and gesture analysis, and the combination of those.
	% He received several Best Paper awards and was nominated Distinguished Researcher by the IEEE Computer Science committee. He received a David Marr Prize, and a Koenderink and a U.V. Helava award. He is a co-founder of 10 spin-off companies. He is a member of the IEEE.
\end{IEEEbiography}


%\vfill

%\input{6appendix}

\end{document}